\documentclass[10pt,twocolumn,letterpaper]{article}

\usepackage{cvpr}      % To produce the REVIEW version
\usepackage{graphicx}
\usepackage{amsmath}
\usepackage{amssymb}
\usepackage{booktabs}
\usepackage{xcolor}
\usepackage{multirow}
\usepackage{hhline}
\usepackage[normalem]{ulem}

\usepackage{placeins}
\usepackage{float}
\usepackage{rotating}

\usepackage[pagebackref,breaklinks,colorlinks]{hyperref}
\usepackage{adjustbox}
\usepackage{amsmath,amsfonts,bm}

\def\eqref#1{equation~\ref{#1}}
\def\1{\bm{1}}

\def\vtheta{{\bm{\theta}}}
\def\va{{\bm{a}}}

\def\vg{{\bm{g}}}

\def\vs{{\bm{s}}}

\def\vy{{\bm{y}}}

\def\mA{{\bm{A}}}

\def\mD{{\bm{D}}}

\def\mF{{\bm{F}}}
\def\mG{{\bm{G}}}

\def\mL{{\bm{L}}}

\def\mS{{\bm{S}}}

\def\mU{{\bm{U}}}

\def\mY{{\bm{Y}}}

\DeclareMathAlphabet{\mathsfit}{\encodingdefault}{\sfdefault}{m}{sl}
\SetMathAlphabet{\mathsfit}{bold}{\encodingdefault}{\sfdefault}{bx}{n}

\usepackage[capitalize]{cleveref}
\crefname{section}{Sec.}{Secs.}
\Crefname{section}{Section}{Sections}
\Crefname{table}{Table}{Tables}
\crefname{table}{Tab.}{Tabs.}

\def\cvprPaperID{10619} % *** Enter the CVPR Paper ID here
\def\confName{CVPR}
\def\confYear{2022}

\begin{document}

%%%%%%%%% TITLE
\title{Learning Graph Regularisation for Guided Super-Resolution\vspace{-0.2em}}

\author{Riccardo de Lutio$^{1,}$\thanks{Equal contribution.}\;\quad Alexander Becker$^{1,\ast}$\;\quad Stefano D'Aronco$^{1}$
\\
Stefania Russo$^{1}$ \;\quad Jan D. Wegner$^{1,2}$ \;\quad Konrad Schindler$^{1}$
 \vspace{.2em} \\
$^{1}$EcoVision Lab, Photogrammetry and Remote Sensing, ETH Zurich
\\
$^{2}$Institute for Computational Science, University of Zurich
\\
{\tt\small firstname.lastname@geod.baug.ethz.ch}
}

\maketitle

%%%%%%%%% ABSTRACT
\begin{abstract}
We introduce a novel formulation for guided super-resolution. Its core is a differentiable optimisation layer that operates on a learned affinity graph. The learned graph potentials make it possible to leverage rich contextual information from the guide image, while the explicit graph optimisation within the architecture guarantees rigorous fidelity of the high-resolution target to the low-resolution source. 
With the decision to employ the source as a constraint rather than only as an input to the prediction, our method differs from state-of-the-art deep architectures for guided super-resolution, which produce targets that, when downsampled, will only approximately reproduce the source. This is not only theoretically appealing, but also produces crisper, more natural-looking images. 
A key property of our method is that, although the graph connectivity is restricted to the pixel lattice, the associated edge potentials are learned with a deep feature extractor and can encode rich context information over large receptive fields. By taking advantage of the sparse graph connectivity, it becomes possible to propagate gradients through the optimisation layer and learn the edge potentials from data.
We extensively evaluate our method on several datasets, and consistently outperform recent baselines in terms of quantitative reconstruction errors, while also delivering visually sharper outputs. Moreover, we demonstrate that our method generalises particularly well to new datasets not seen during training.
\end{abstract}

%%%%%%%%% MAIN BODY

\section{Introduction}
\label{sec:intro}

Guided super-resolution takes as input two images of different resolution, a low-resolution \textit{source} and a high-resolution \textit{guide} from a different domain. It returns a high-resolution version of the source as output, termed the \textit{target}. This task is relevant in many practical applications such as medical~\cite{zhang2018longitudinally} and satellite imaging~\cite{Lanaras2018}, where performing a diagnosis or analysis from low-quality images can be extremely difficult. 
Another very popular example in computer vision is upsampling depth maps, where the low-resolution depth is the source, a conventional grayscale or RGB image is the guide, and the target is a high-resolution depth map.
Consumer-grade depth sensors provide low-resolution depth maps, but a high-resolution RGB camera is usually mounted on the same device and can acquire a high-resolution image of the same scene.
\begin{figure}[t]
  \centering
  \includegraphics[width=.9\linewidth]{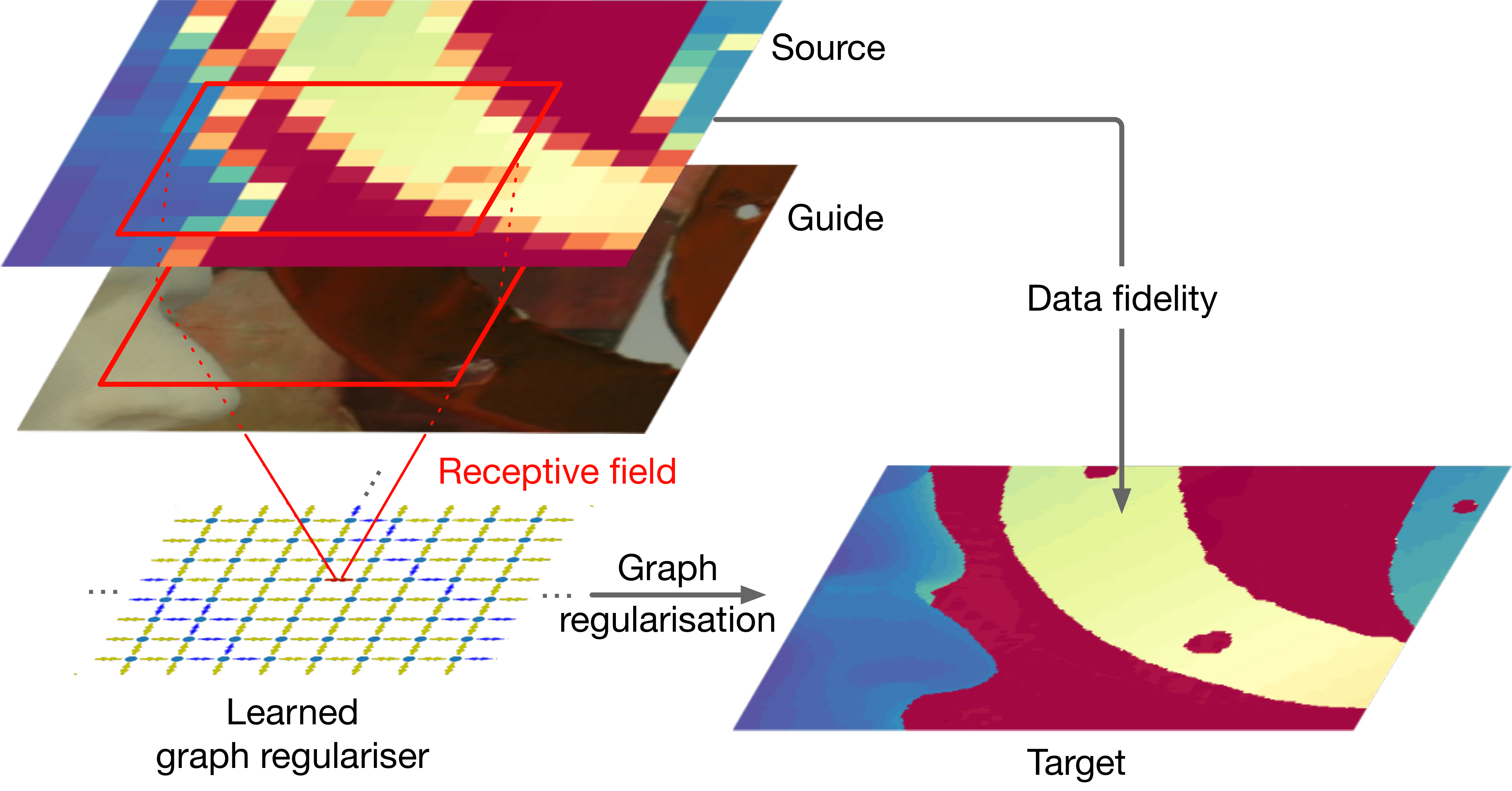}
   \caption{Our method takes as input a low-resolution \textit{source} image and a high-resolution \textit{guide} image of another modality to build a graph using high-level image features. The graph is then used in a differentiable optimisation layer as a regularisation to reconstruct the \textit{target}.}
   \label{fig:teaser}
\end{figure}
Guided super-resolution methods can be divided into two main categories, conventional and deep learning-based methods. The former typically cast the task into an optimisation problem~\cite{Ferstl2013, dong2016color, Diebel2006, Ham2018}. The goal is to create a high-resolution target image that, when downsampled, matches the source, while at the same time complying with an appropriate regularisation term that favours a desired image characterstic such as (piecewise) smoothness. Deep learning methods~\cite{kim2016accurate, Tak-Wai2016, he2021, Kim2021 , Sun2021cvpr} instead rely on a dataset of source/guide/target triplets to learn a feed-forward mapping from the source and guide to the target.
To that end the model must learn the statistical correlations that allow it to transfer high-frequency details from the guide to the target, while, at the same time, ensuring that the predicted target stays close to the source.
A considerable advantage of the conventional approach is that, by individually solving a properly formulated optimisation for each image, the prediction is usually guaranteed to match the source. On the other hand, designing an adequate regularisation term based on low-level image features is a complicated task.
Deep learning methods exhibit rather complementary strengths: as long as one has access to enough training data and that data is representative of the images encountered at test time, these methods tend to perform very well, due to the unmatched ability of deep networks to mine complex, highly informative features from images. On the other hand, with limited training data, or when there is a domain shift between the training and the test set~\cite{dataset_bias}, feed-forward methods can no longer guarantee that downsampling the predicted target will produce the source, thus contradicting the fundamental relation behind super-resolution.

In this work, we show how to combine the two schools, and learn the graph of an optimisation-based super-resolution scheme,\,--\,see Figure \ref{fig:teaser}.
In particular, we learn a mapping from the two inputs (source and guide) to the edge potentials (also called edge weights) of an affinity graph between pixels of the target. The learned graph serves as the regulariser for an optimisation-based reconstruction of the high-resolution target, which is particularly suited for signals with a piecewise smooth structure.
This entire mapping is trained end-to-end: the mapping function, which is parametrised as a convolutional network, is learned from training data, by back-propagating the gradients of the loss through the optimisation layer.
CRF-RNN \cite{zheng2015} also proposed to perform an online optimisation and include a graph in their network for semantic segmentation. However, they construct a dense graph and use an RNN to approximate the inference of the posterior. In contrast, we show that a sparse, local graph is sufficient while performing exact maximum a posteriori inference.
We test our method on three different guided depth super-resolution datasets and show that it compares favourably against conventional and deep learning-based methods, across a range of upsampling factors from  $\times4$ to $\times16$. We further show that our proposed method is much more robust to distribution shifts and can effectively generalise across datasets.

In summary, the contributions of this paper are the following: \emph{(i)} We introduce a novel formulation of guided super-resolution, where a deep feature extractor is trained to derive the edge potentials for a graph-based energy minimisation from the input (source and guide) images; \emph{(ii)} we develop a differentiable optimiser for the graph regularisation, taking advantage of the sparse graph connectivity to efficiently process large input patches up to $256^2$ pixels\footnote{Code is available at \url{https://github.com/prs-eth/graph-super-resolution}}; \emph{(iii)} in this way, our scheme therefore combines the power of learned, deep feature extractors with large receptive fields and the rigor of graph-based optimisation in an end-to-end trainable framework. As a result, it produces crisp, natural-looking images that correctly adhere to the underlying image formation model.
\section{Related Work}
\label{sec:related}

At a conceptual level, guided super-resolution can be seen as a form of guided filtering~\cite{He2013}, where the source image is first naively upsampled to match the target resolution, and then enhanced with some transformation that is guided by the local structure of both the (upsampled) source and the guide.

\subsection{Optimisation Methods}

\textit{Local optimisation} methods are variants of the filtering procedure described above. Here, the source is first upsampled, then a local filter controlled by the values of the guide \cite{Kopf2007,Yang2007} is applied to it. Extensions of these methods include the use of geodesic distances to define the filter \cite{Liu2013}, or constructing it by combining the contrast in both the guide and source images \cite{Chan2008}.

\textit{Global optimisation} methods construct a global energy function over all pixels and minimise it to obtain the target. The energy generally consists of two parts: a data fidelity term that ensures that the target stays close to the source, and a prior term that regularises the otherwise ill-posed problem of super-resolution.
Data fidelity is typically defined as a distance term between the source and the downsampled target. The regulariser, in the guided setting, is not an isotropic smoothing but it is modulated by the guide. Depending on the parametrisation, the global energy minimisation can be viewed as Markov Random Field (MRF) inference \cite{Diebel2006}, as a form of non-local means \cite{Park2011}, or as a variational inference with an anisotropic version of total generalised variation (TGV) \cite{Ferstl2013}. Some works \cite{Yang2012,Yang2014} have also proposed to replace the TGV prior with an auto-regressive model.
%
%Another school embeds the idea of guided filtering in a global optimisation framework (rather than local filtering). 
The fast bilateral solver \cite{Barron2016} solves a sparse linear system \cite{Barron2015} to obtain bilateral-smooth outputs with sharp discontinuities. The SD filter \cite{Ham2018} formulates guided image filtering as a non-convex optimisation problem that exploits static and dynamic guidance. The Pixtransform method \cite{deLutio2019} estimates a mapping from guide to target individually for each pixel and spatially smoothens the mapping function, rather than the target output. In a similar spirit, \cite{Pan2019} predicts the target as a linear function of the guide, with coefficients that vary spatially, modulated by the guide and the source.
The Guided Deep Decoder (GDD) \cite{Uezato2020} adapts the deep image prior \cite{Ulyanov2018} to guided super-resolution of hyper-spectral images. A random noise map is decoded into a target that has maximal data fidelity to the source, guided by feature maps obtained by a joint encoder-decoder branch for the guide.
Cross-Modality Super-Resolution (CMSR) \cite{Shacht_2021_CVPR} also fits a neural architecture to the individual source/guide pair, allowing to optimise for individual alignment errors. 
%%%%The network is trained at a reduced resolution, where the source can serve as high-resolution supervision, assuming scale-invariance of the mapping.
%\ks{These last two, GDD and CMSR, should go under optimisation methods, isn't it? The computational architecture may be a network, but if you fit it only to the two inputs, you optimise (train) at inference time, and learn nothing from other data, so it is just like our Pixtransform.}
%
\cite{PARK201950} proposes to (over-)segment the images and attribute planar disparities to each super-pixel, while smoothness of the disparities across super-pixels is encouraged by connecting them into an MRF.

In \cite{Rossi2020}, the authors also propose to build a graph to encourage smoothness of the target in regions where the guide is also smooth. Contrary to our work, their graph is based on raw colour differences (similar to  \cite{Diebel2006,PARK201950}) whereas our graph encodes affinity between deep, latent features, derived not only from the guide but also from the source, and trained in an end-to-end fashion to optimally support the super-resolution task.

\subsection{Learning Methods}

The other large family of guided super-resolution methods are learning-based. Following a general trend towards supervised machine learning, the hope is that one can outperform conventional models by learning from data how to best fuse the source and the guide to recover the target.
%
%%%The price to pay is that one must have access to large datasets with known reference (i.e., high-resolution targets), assuming they are representative of the unseen test data.
%
Perhaps the first learning-based methods for guided super-resolution were those learning dictionaries of source, guide and target patches. At test time, the source and guide were then (soft-)matched to the dictionary to retrieve suitable target patches and assemble the target image \cite{Li2012,Kwon2015}.

More recently, deep learning methods have become predominant for guided super-resolution. These approaches work by parametrising the non-linear mapping from the two inputs\,---\,guide and source\,---\,to the target as a convolutional neural network, and learning its weights directly.
The deep joint image filter \cite{Li2016,Li2019} feeds the upsampled source and the guide directly into a standard encoder-decoder architecture. The deep primal-dual network \cite{Riegler2016} follows a similar strategy, but outputs a residual correction to the naively upsampled source. Additionally, the output is refined with non-local total variation, unrolled into a sequence of network layers.
The Multi-Scale Guided network (MSG-Net) \cite{Tak-Wai2016} implements a new strategy, to \emph{en}code only the guide, extract rich hierarchical features at different levels of the encoder, and append them to the corresponding levels of a network that \emph{de}codes the source into the target through a final reconstruction layer. This integrated multi-scale guidance from the guide to the upsampled source allows to resolve ambiguity in depth map upsampling.
This design has inspired several other works: PMBANet \cite{Ye2020} adds multi-branch aggregation blocks; the Fast Depth Super-Resolution network (FDSR) \cite{he2021} adds a high-frequency layer to extract fine details from the guide, and strives for a computationally efficient, yet effective design. DepthSR-Net \cite{Guo2019} integrates the idea in a residual U-Net architecture \cite{ronneberger2015u}. %
First, the source is naively upsampled to the desired resolution, then the residuals between this naive interpolation and the corresponding target are learned using the hierarchical features as input pyramid in the encoder structure. %
In \cite{Wen2019}, an explicit coarse-to-fine cascade of networks is used to iteratively refine the output and progressively add high-frequency details. 
In \cite{Sun2021cvpr} two networks are trained collaboratively, one for monocular depth estimation from the guide and one to super-resolve the source. Furthermore, there is an auxiliary structure prediction task to mitigate differences between depth and intensity discontinuities.
Also in a very recent work, \cite{safin2021unpaired} explores learning depth super-resolution from unpaired data, using a learnable degradation model, and surface normal estimates as additional features to obtain more accurate depth maps.

Several authors have experimented with modifications of the basic Convolutional Neural Network (CNN) layers to enable modulation based on the guide. The Pixel-Adaptative Convolutional (PAC) network \cite{su2019} proposes a novel type of learned filters where the convolution is conditioned on other features. For guided super-resolution, these conditioning features are extracted from the guide. 
Channel attention is used in \cite{Song2020l} to improve super-resolution of channels with abundant high-frequency content.
The Deformable Kernel Network (DKN) \cite{Kim2021} applies sparse, spatially-variant kernels to predict a set of neighbours and associated weights for each target pixel, such that their weighted mean yields the pixel's value.
\section{Method}\label{sec:methodology}

\subsection{Notation and Problem Statement}
Throughout, we denote matrices and higher-order tensors with uppercase bold letters $\mA$, and their flattened, 1D vector version with corresponding lowercase letters $\va$. In our guided super-resolution setting we are given a guide $\mG$ with spatial dimensions $H\!\times\!W$ and $C$ channels, as well as a low-resolution source $\mS$ of dimension $h\!\times\!w$.
For simplicity, we will assume that the source has a single channel, as the extension to multiple channels is straight-forward. The ratio between the spatial dimensions of the guide and the source is the upsampling factor $k=H/h=W/w$. The goal is to upsample $\mS$ to a target $\mY$ with the same spatial resolution of $\mG$.
We denote with $\mD$ the \textbf{downsampling operator} that maps $\vy$ to $\vs$. In our case the downsampling is a weighted average over a $k \times k$ window of the image $\mY$ (a \emph{point spread function}). Note that some authors instead assume that $\mS$ is not downsampled, but rather represents a sparsely sampled version of the target $\mY$, which need not be deconvolved. Due to the finite area of pixels on the sensor, respectively the beam divergence in laser-based scanners, this sparse subsampling without low-pass filtering is not a very realistic model for most practical sensing systems.

\begin{figure*}[t]
    \centering
    \includegraphics[width=0.95\textwidth]{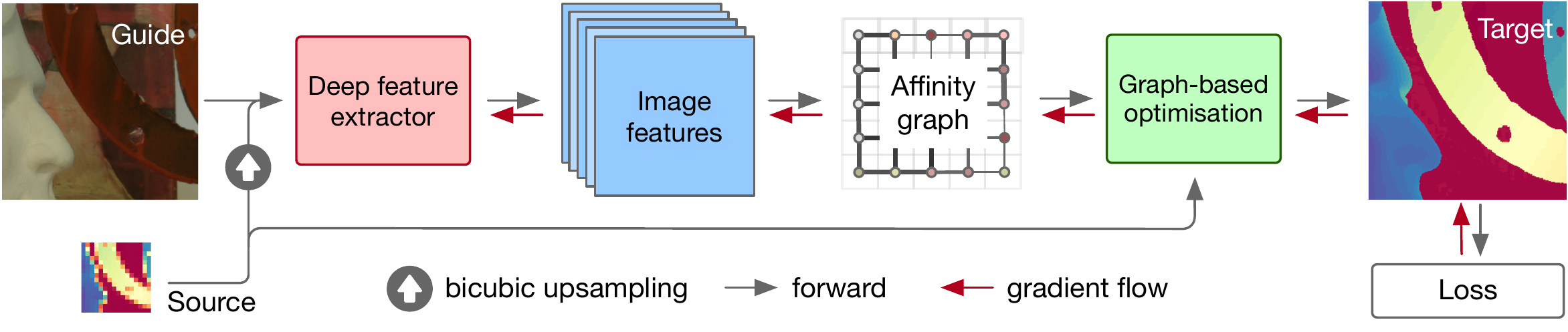}
    \caption{The architecture of our approach. A neural network backbone is employed to  extract deep feature maps from the guide and source images. A graph over pixels is constructed based on pairwise affinities derived from these feature maps. Finally, a quadratic optimisation problem is solved for a target image that is in agreement with both the low-resolution source and the structure of the graph. Crucially, the graph optimisation layer is differentiable and our method is thus end-to-end trainable.}
    \label{fig:model}
\end{figure*}

\subsection{Graph Regularisation}
\label{sec:graph_reg}

A natural way to formalise the guided super-resolution problem mathematically is as an energy minimisation:
\begin{equation}
 \underset{\vy}{\text{argmin}} \quad f\big(\mD\vy,\vs\big) + \lambda\cdot r(\vy),
\end{equation}
where $f$ is the data fidelity term that measures how well the downsampled target matches the source, and $r$ is a prior, respectively regulariser for the reconstructed target, and $\lambda$ a parameter that weights the effect of the regularisation. 

The \textbf{data fidelity term} serves to ensure similarity between $\mD\vy$ and the source $\vs$, typically in the form of an $l_1$ or (squared) $l_2$ norm. In this work we use the latter, $f(\mD\vy,\vs) = \|\mD\vy - \vs\|_2^ 2$.

An effective \textbf{regulariser} that has often been used successfully for images~\cite{Diebel2006, Rossi2020} is to encourage smoothness of the reconstructed signal \wrt some graph defined over the image pixels.
The affinity matrix of that graph is denoted by $\mA$ and has size $HW \times HW$. It describes which pixels are connected, i.e., have direct, first-order influence on each other. The generic element $A_{ij}$ represents the weight of the edge connecting pixel $i$ to pixel $j$, for all pairs of pixels that are not directly connected, $A_{ij}=0$. The degree matrix $\mU$ is a diagonal matrix with entries constructed by summing the weights of all edges that meet at a node, $U_{ii} = \sum_j A_{ij}$. Finally, the graph Laplacian $\mL$ is defined as $\mL = \mU - \mA$. For a signal defined on the graph nodes\,--\,in our case the image\,--\,the quantity $\vy^T \mL \vy $ is a measure of how smooth that signal is on the graph. Encouraging smoothness is an effective regulariser, provided that the graph matches the intrinsic structure of the signal. The objective then becomes:
\begin{equation}
 \underset{\vy}{\text{argmin}} \quad \|\mD\vy - \vs\|_2^ 2 + \lambda \vy^T \mL \vy\;.
 \label{eq:graph_reg_opt}
\end{equation}
What remains is to construct the right graph, i.e., to determine the ``natural" connectivity between the pixels of the target $\mY$. This is not trivial, but in guided super-resolution we can leverage the guide $\vg$, which shares the same, high resolution with $\mY$. The graph thus becomes a function $\mL(\vg)$ of the guide. Note that this does not imply constructing the graph from the guide's raw brightness (resp., contrast) values. Rather, one may as well derive it from more abstract per-pixel features. As we will show, a particularly useful procedure is to learn those features from the data, such that the graph is optimally adapted to the specific super-resolution task at hand. 

In order to tailor the graph structure of the regulariser to the problem, we feed both the source and the guide through a CNN, to obtain a deep feature representation $\mF=f_\vtheta(\mG,\mS)$ of size $H\!\times\!W\!\times\!M$, with $M$ the channel depth of the representation and $\vtheta$ the trainable parameters of the network.
For efficiency, we restrict the graph to have fixed topology, where each pixel is connected (at most) to its 4-neighbours in the 2D pixel lattice. Longer-range connectivities are in principle possible, but greatly increase the computational effort, with rapidly diminishing returns. 
Indeed, our setup attaches deep features to the graph nodes, which encode a large receptive field and capture semantic and long-range information in the guide image beyond the 4-neighbour topology. 
The weights of the graph edges are defined as standard negative exponential affinities between the learned features:
\begin{equation}
A_{ij} = e^{-\frac{\|F_{i} - F_{j}\|_2^2}{M\mu}},
\end{equation}
where $\mu$ is a learnable scaling parameter.

\subsection{Optimisation Layer}
\label{sec:optim_layer}
Let $\vy^\star$ denote the minimiser of Eq.~(\ref{eq:graph_reg_opt}), and $\vy_{\text{gt}}$ the ground truth target values of some training set. We can then assemble triplets $(\vg,\vs,\vy_{\text{gt}})$ and optimise the graph construction to minimise the loss between $\vy^\star$ and $\vy_{\text{gt}}$:
\begin{equation}
\vtheta*=
\underset{\vtheta}{\text{argmin}} \quad \mathbb{E}_{p(\vg,\vs,\vy_{\text{gt}})} \big[ l\big(\vy^\star(\vtheta),\vy_{\text{gt}}\big) \big]\;,
 \label{eq:graph_training}
\end{equation}
where $l$ is an appropriate loss function, for instance an $l_1$ loss or a Mean Squared Error (MSE) loss.

This means that, in order to train the feature extractor, we must compute the gradient of the loss in Eq.~(\ref{eq:graph_training}) \wrt the graph.
Towards that end, we first notice that Eq.~(\ref{eq:graph_reg_opt}) is a quadratic problem, and equivalent to solving the linear equation system:
\begin{equation}
\big(\lambda \mL(\vtheta) + \mD^T\mD\big) \vy^\star = \mD^T\vs,
\end{equation}
here we have made it explicit that the graph Laplacian $\mL$ is the only term that depends on the network parameters $\vtheta$. For error backpropagation we must map the gradient $\frac{\partial l}{\partial\vy^\star}$ \wrt the reconstructed image to the entries of $\mL$.
Using the implicit function theorem~\cite{Barron2016} we obtain:
\begin{equation}
\tfrac{\partial l}{\partial \mL} = - \lambda \tfrac{\partial l}{\partial \mD^T\vs} {\vy^\star}^T, \;
(\lambda \mL(\vtheta) + \mD^T\mD) \tfrac{\partial l}{\partial \mD^T\vs} =  \tfrac{\partial l}{\partial\vy^\star}
\end{equation}
In order to backpropagate the loss, we must solve a second linear equation system, which then yields the derivatives for individual entries of the graph Laplacian.
%
%Note that the linear system can be solved either with direct methods or by implementing for instance a conjugate gradient. 
Note that the derivative \wrt the Laplacian is a dense matrix, which is impractical (e.g., for an image of $256^2$ pixels this matrix has $\approx4$ billion elements). Fortunately, we can exploit the fact that the graph topology is fixed and compute the gradient only \wrt non-zero entries of $\mL$ (i.e., index pairs of the 4-neighbourhood). Once the gradients for the graph weights have been computed, they are propagated through the deep feature extractor.

Finally, we summarise our proposed model, see Figure~\ref{fig:model}. The feature extractor $f_\vtheta(\mG,\mS)$ computes the deep features from the guide and source images, and these features inform the weights of a 4-neighbour graph. 
The graph, together with the source $\mS$, forms the input to the optimisation problem of Eq.~(\ref{eq:graph_reg_opt}) that estimates the target. During training, a loss computed between the prediction and the ground truth steers the feature extraction such that the graph weights optimally regularise the prediction of the high-resolution target. Note that at test time we must solve a quadratic problem in order to predict the target image. To do so, very efficient algorithms are available, although it is of course not as fast as a conventional forward pass.

\section{Experimental Results}
\label{sec:experiments}

In this section we describe the evaluation of our proposed method on the task of RGB-guided depth map super-resolution. We conduct our experiments on three widely used RGB-D datasets. For each dataset, we compare our approach to several guided super-resolution baselines.
All algorithms are evaluated at 3 upsampling factors -- $\times 4$, $\times 8$ and $\times 16$ -- on the following datasets:

\textbf{Middlebury} \cite{Scharstein2007,Scharstein2001,Scharstein2014,Scharstein2003,Hirschmuller2007} We use all 50 RGB-D images available from the Middlebury 2005-2014 datasets. We split the data randomly into 40 images for training, 5 for validation and 5 for testing. A challenging aspect of this dataset is that it contains missing values in the depth ground truth. For generating the source, we therefore only take into account valid pixels during downsampling. Furthermore, we generate a pixel validity mask for both the target and source, so we can ignore the invalid pixels during training and testing.

\textbf{NYUv2} \cite{Silberman2012} consists of 1449 RGB-D images captured with a Microsoft Kinect. We randomly split these into 849 images for training, 300 for validation and 300 for testing.

\textbf{DIML} \cite{DIML1,DIML2,DIML3,DIML4} is a large-scale dataset comprised of 2M RGB-D frames in total. For our evaluation, we use the high-resolution indoor sample subset, which was acquired using a Microsoft Kinect. From this data we construct a split of 1440 images for training, 169 images for validation and 503 images for testing.

We compare our model to the \textit{Guided Filter} (GF) \cite{He2013}, the \textit{Static/Dynamic filter} (SD) \cite{Ham2018}, the \textit{Pixtransform} \cite{deLutio2019}, the \textit{MSG-Net} \cite{Tak-Wai2016}, the \textit{Deformable Kernel Network} (DKN) and its fast version (FDKN) \cite{Kim2021}, the PMBANet \cite{Ye2020}, and finally to the \textit{Fast Depth Super-Resolution} (FDSR) \cite{he2021}. We were not able to compare to the recent work by \cite{Sun2021cvpr} since no code has been released at the time of writing. For all other methods, we use the respective publicly available code. We implemented our method using PyTorch \cite{Paszke2019}. The graph-based optimisation layer is realised using the sparse matrix support of the CuPy library \cite{Okuta2017}, allowing for a GPU-accelerated implementation of the forward and backward pass. In order to solve the linear systems of equations needed in the optimisation, we implement the conjugate gradient method~\cite{numerical_opt}.
We use a U-Net \cite{ronneberger2015u} network with a ResNet-50 \cite{he2016deep} encoder pretrained on ImageNet~\cite{deng2009imagenet} as feature extractor for the graph weights prediction. We also use a simple gradient clipping as it improves the stability of the training procedure.
As baseline we further compare to a version of the proposed method where the RGB features of the guide and the bicubic upsampling of the source are the only pixel features used to construct the graph, i.e., no deep feature extractor is used.

We train all learned methods using the Adam \cite{Kingma2014} optimiser. Varying with the specific dataset, we fix the same batch size, initial learning rate and scheduling strategy for all methods. For fairness of comparison, we trained all learned methods with both our configuration for these hyperparameters and their original ones (when indicated) and report the best results. The detailed hyperparameter settings for each dataset and method as well as additional experimental results can be found in the supplementary material. For all learned methods we further make use of data augmentations during training, consisting of random cropping, random horizontal flipping and random rotation, where the rotation angle is sampled from $\mathcal{U}(-15^{\circ},15^{\circ})$. All methods are evaluated on patches of $256^2$ pixels, however, for some methods training on such large patches was infeasible due to memory constraints, in which case we used patches of size $128^2$ or even $64^2$ (for PMBANet with a factor $\times4$).

\label{sec:training}

\subsection{Learning Graph Weights}

\begin{figure}[t]
    \centering
    \includegraphics[width=0.9\linewidth]{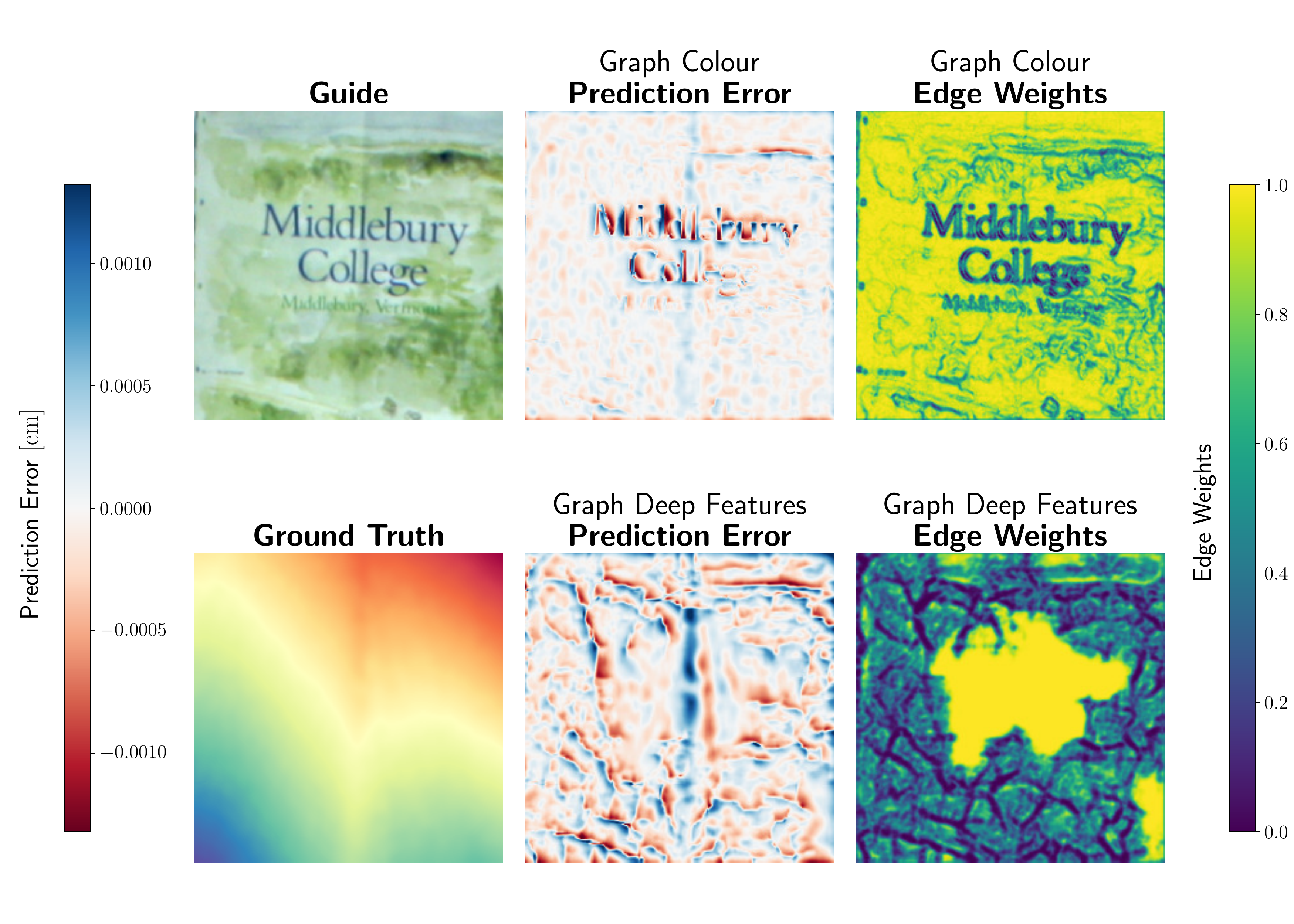}
    \includegraphics[width=0.9\linewidth]{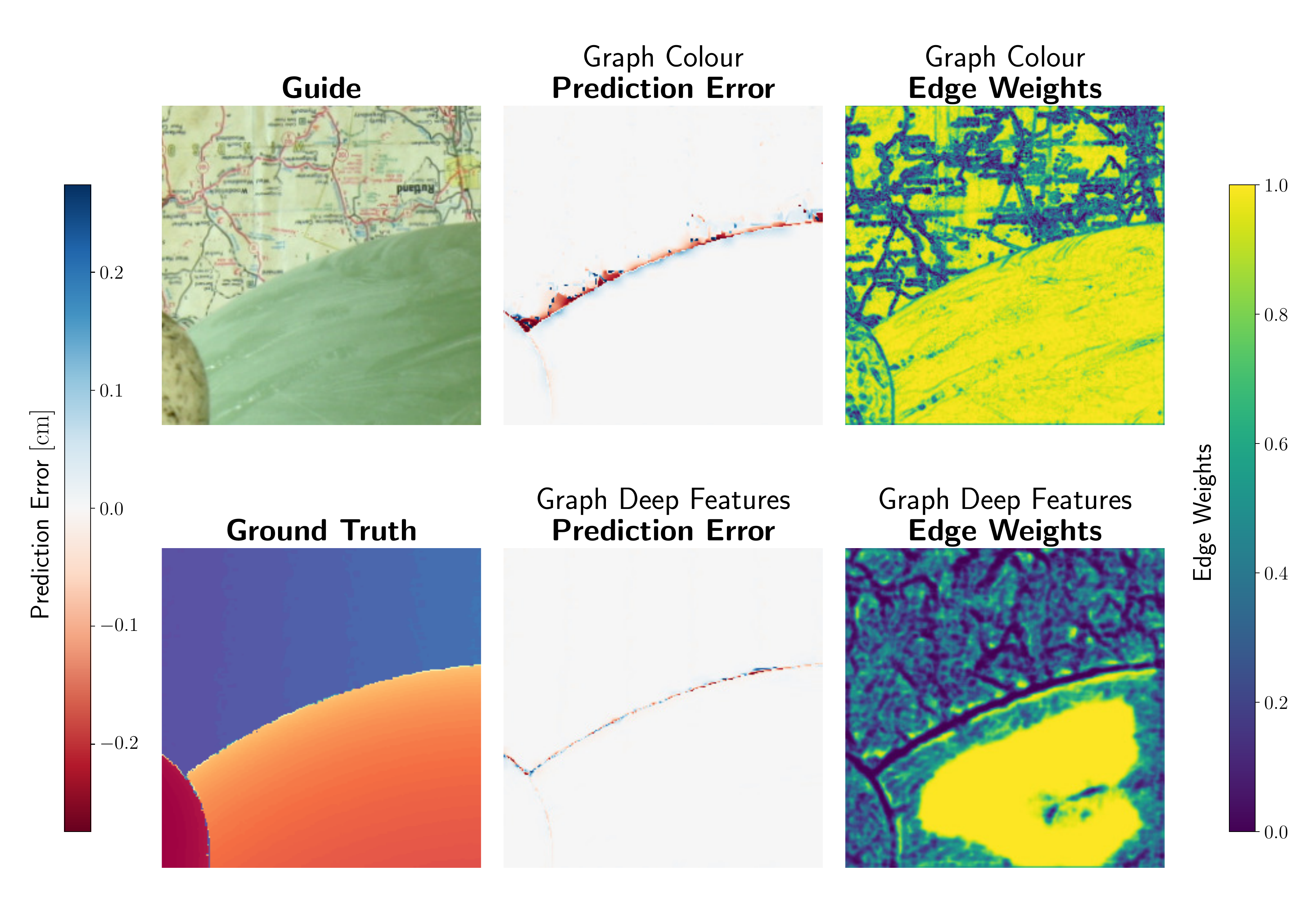}
    \caption{Importance of learned edge potentials. We visualise the total affinity of each pixel to its four neighbours when derived from raw colour (top) or from deep features (bottom). Examples are from the Middlebury test set. 
    }
    \label{fig:weights}
\end{figure}
As previously mentioned, we believe that smoothing based on a graph that is defined on four \emph{local} neighbours only is adequate for the problem at hand, as long as the features used to create the graph encode a sufficiently large context. 
This is of course the case when using deep features that are learned from an entire dataset using a CNN with a large receptive field. In Figure~\ref{fig:weights}, we compare the graph obtained from plain colour information to the graph obtained from learned deep features for two selected examples from the Middlebury test set.
The graph is visualised by displaying, for each pixel, the sum of the four edges that connect it to its neighbours. Areas where the connections of the nodes are strong appear in yellow, and in these areas the graph regularisation will enforce smoothness in the predicted target. On the other hand, in areas where the edge weights are small, the graph smoothness term is weak (it disappears as weights approach zero) and the target is allowed to reveal depth discontinuities.
Figure~\ref{fig:weights} demonstrates that the proposed method is able to extract semantical information from the guide image and to transfer it to the edge potentials.
In the top example, the model has learned that a high-contrast text is part of the surrounding object, thus it predicts high edge weights for the respective image area and effectively enforces smooth depth. This correctly results in the text not being carried over to the target, an effect that is observable in the prediction based on colour features.
On the other hand, the bottom image shows that our model has learned to detect object boundaries and highlights them even if both background and foreground have very similar colours.
In fact, the learned edge weights are much lower around the depth discontinuities compared to the colour-based weights, which implies that the learned graph is able to recreate a sharp edge in the target prediction. In contrast, the graph built from colour information cannot perform a proper cut, leading to bleeding artifacts.

\begin{table*}[t]
\centering
\resizebox{\textwidth}{!}{%
\begin{tabular}{c c ccccccccccc}
& & &  GF \cite{He2013}   & SD filter \cite{Ham2018} & Pixtransform \cite{deLutio2019} & MSG-Net \cite{Tak-Wai2016} & DKN \cite{Kim2021}  & FDKN \cite{Kim2021} & PMBANet \cite{Ye2020} & FDSR \cite{he2021} & Ours - Colour & Ours \\ \hline \hline
 
 \multirow{6}{*}{\rotatebox{90}{Middlebury}} & \multirow{2}{*}{$\times 4$}  & MSE & 33.3 & 24.9 & 39.8 & 4.13 & 4.29 & 3.60 & 4.72 & 7.72 & 14.8 & \textbf{3.04} \\ %\cline{3-13} 
                                             &                              & MAE & 1.27 & 0.46 & 0.79 & 0.22 & 0.18 & 0.16 & 0.25 & 0.35 & 0.42 & \textbf{0.13} \\ \cline{2-13}

                             &   \multirow{2}{*}{$\times 8$}  & MSE & 40.5    & 82.5 &  32.7 & 10.5 & 11.2 & 10.4 & 9.48 & 23.2 & 68.3 & \textbf{7.26} \\ %\cline{3-13} 
                              &                               & MAE  & 1.49 &    0.86 &  0.82 & 0.43 & 0.38  & 0.37 & 0.38 &  0.69 & 0.83 & \textbf{0.24} \\  \cline{2-13}

                               & \multirow{2}{*}{$\times 16$} & MSE & 67.4 &   511 & 41.5 & 34.2 & 47.6 & 38.5 & 30.6 & 55.4 & 297 & \textbf{24.7} \\ %\cline{3-13} 
                                &                             & MAE  & 2.21 &   1.73 & 1.24 & 1.06 & 1.42  &  1.18 & 0.89 & 1.51 & 1.69 & \textbf{0.67} \\ \hline\hline

\multirow{6}{*}{\rotatebox{90}{NYUv2}} &        \multirow{2}{*}{$\times 4$}  & MSE & 114  &   36.0 & 112 & 6.85 & 11.4 & 8.07 & 10.8 & 10.5 & 19.0 & \textbf{6.45} \\ %\cline{3-13} 
                            &                                 & MAE  & 3.91  &   1.31 & 3.61 & 0.81 & 1.03 & 0.85 & 0.93 & 0.94 & 1.11 & \textbf{0.73}\\ \cline{2-13}

                             &   \multirow{2}{*}{$\times 8$}  & MSE & 142  &   105 & 122 & 24.1 & 29.8 & 29.9 & 31.5 & 35.4 & 68.4 & \textbf{19.6} \\ %\cline{3-13} 
                              &                               & MAE  & 4.47  &   2.57 & 3.86 & 1.66 & 1.82 & 1.80 & 1.79 & 1.96 & 2.30 & \textbf{1.42} \\  \cline{2-13} 

                               & \multirow{2}{*}{$\times 16$} & MSE & 249  &    533  & 219 & 84.5 & 115 & 113 & 84.9 & 179 & 264 & \textbf{67.5} \\ %\cline{3-13} 
                                &                             & MAE  & 6.34  &    5.07 & 5.40 & 3.35 & 4.01  & 3.95 & 3.26 & 4.68 & 4.56 & \textbf{2.90} \\ \hline\hline

\multirow{6}{*}{\rotatebox{90}{DIML}}&     \multirow{2}{*}{$\times 4$}  & MSE & 25.6  & 10.5 & 20.7 & 1.73 & 3.47 & 2.2 & 3.05 & 2.75 & 7.02 & \textbf{1.68}  \\ %\cline{3-13} 
                                &                        & MAE  & 1.45 &  0.40 & 1.15 & 0.22 & 0.33 & 0.23 & 0.31 & 0.29 & 0.35 & \textbf{0.20} \\ \cline{2-13} 
    
                        &    \multirow{2}{*}{$\times 8$}  & MSE & 34.1 &  44.9 & 23.0 & 4.13 & 5.47 & 5.95 & 5.87 & 8.40 & 15.2 & \textbf{3.51} \\ %\cline{3-13} 
                         &                                & MAE  & 1.77 & 0.83 & 1.26 & 0.40 & 0.45 & 0.47 & 0.47 & 0.66 & 0.67 & \textbf{0.31} \\ \cline{2-13}

                          &  \multirow{2}{*}{$\times 16$}  & MSE & 66.3  & 411 & 39.3 & 13.0 & 19.3 & 20.8 & 13.8 & 32.9 & 133 & \textbf{9.45} \\ %\cline{3-13} 
                           &                              & MAE  & 2.74  & 1.91 & 1.78 & 0.93 & 1.20 & 1.24 & 0.87 & 1.66 & 1.72 & \textbf{0.68} \\ \hline\hline

\end{tabular}%
}
\caption{Performance comparison with the state-of-the-art algorithms on the Middlebury \cite{Scharstein2007,Scharstein2001,Scharstein2014,Scharstein2003,Hirschmuller2007}, NYUv2 \cite{Shelhamer2017} and DIML \cite{DIML1,DIML2,DIML3,DIML4} datasets for different values of upsampling factors. The table shows the MSE (in $\text{cm}^ 2$) and MAE (in cm).}
\label{tab:cmp}
\end{table*}

\begin{table*}[t]
\centering
\resizebox{\textwidth}{!}{%
\begin{tabular}{lcc ccc cccc cc}
 Testing Dataset & &  GF \cite{He2013} & SD filter \cite{Ham2018} & Pixtransform \cite{deLutio2019} & MSG-Net \cite{Tak-Wai2016} & FDKN \cite{Kim2021}  & PMBANet \cite{Ye2020} & FDSR \cite{he2021} & Ours - Colour & Ours \\ \hline \hline

\multirow{2}{*}{DIML}  & MSE & 34.1 &  44.9 & 23.0 & 5.76 & 6.74 & 7.35 & 7.73 & 20.5 & \textbf{4.95} \\ %\cline{2-11} 
                       & MAE  & 1.77 &  0.83 & 1.26 & 0.51 & 0.53 & 0.59 & 0.74 & 0.77 & \textbf{0.40} \\
                       & MSE (low-resolution) & 17.7 & 1.45 & 6.19 & 6.16 & 0.20 & 0.04 & 0.45 & 0.03 & $\mathbf{2.4 \cdot 10^{-3}}$
                       \\ \hline\hline

\multirow{2}{*}{Middlebury}  & MSE & 40.5  & 82.5 & 32.7 & 11.0 & 10.0 & 9.62 & 18.4 & 23.9 & \textbf{8.25} \\ %\cline{2-11} 
                             & MAE  & 1.49  & 0.86 & 0.82 & 0.54 & 0.43 & 0.46 & 0.73 & 0.91 & \textbf{0.35} \\
                             & MSE (low-resolution) & 17.9 & 1.86 &  22.5 & 5.01 & 0.20 & 0.06 & 7.20 & 0.08 & $\mathbf{1.1 \cdot 10^{-3}}$
                             \\ \hline\hline

\end{tabular}%
}
\caption{Performance comparison with the state-of-the-art algorithms on cross-dataset generalisation. All learned methods have been trained on the NYUv2 dataset \cite{Shelhamer2017}. The table shows the performance of the methods on the DIML \cite{DIML1,DIML2,DIML3,DIML4} and Middlebury \cite{Scharstein2007,Scharstein2001,Scharstein2014,Scharstein2003,Hirschmuller2007} datasets for a $\times 8$ upsampling factor. The table shows the MSE (in $\text{cm}^ 2$), MAE (in cm) and low-resolution MSE (in $\text{cm}^2$).
}
\label{tab:cmp_crosstesting}
\end{table*}

\subsection{Depth Super-Resolution Evaluation}
In Table \ref{tab:cmp} and Figure \ref{fig:depth_upsampling} we quantitatively and qualitatively compare our method to all selected baselines for the Middlebury, NYUv2 and DIML datasets. 
We outperform all other methods \wrt both MSE and MAE metrics for upsampling factors of $\times4$, $\times8$ and $\times16$. 
It is observable from the table that the tested methods perform rather differently among the three datasets. Conventional methods tend to perform generally worse than the learned ones. In particular, Pixtransform~\cite{deLutio2019} shows a rather flat performance curve with mediocre performance on low upsampling factors but also no abrupt performance drop for higher upsampling factors. In terms of visual results, the method reveals many artifacts being carried over from the guide. The SD filter~\cite{Ham2018} instead has good performance on MAE but MSE performance degrades fast for larger upsampling factors. Visually it captures some edges very well, whereas it completely misses and smoothes out others, as seen in Figure \ref{fig:depth_upsampling}.
FDKN and DKN \cite{Kim2021} attain worse performance than expected across the datasets, especially quantitatively.
It appears that these methods are tuned to sparsely downsampled source images and not well suited for realistic (not impulse-shaped) point spread functions.
Our method by contrast achieves good quantitative performance across all three datasets, while producing visually crisp images. It is particularly effective at larger upsampling factors, showing the advantages of a hybrid model that leverages a deep learning backbone alongside a conventional online optimisation layer. Finally, it is more robust to domain shifts, as we explain in the next paragraph.

\paragraph{Cross-dataset generalisation.} A major advantage of our work is that the prediction, after downsampling, is constrained to match the source. This additional constraint affords the model better robustness against domain shifts between training and testing (at the cost of added computation at inference time). To quantify this behaviour, we perform a cross-dataset generalisation experiment. For all methods, we train on NYUv2 and test the resulting model on  Middlebury and DIML. As can be seen in Table~\ref{tab:cmp_crosstesting}, we outperform all other methods by a significant margin. Furthermore, our prediction matches the source almost perfectly when downsampled, as measured by the low-resolution MSE.

\newcommand\FigsGridWidth{0.09}
\newcommand\FigsGridVspace{0.05cm}
\newcommand\FigsGridVspacee{0.15cm}
\newcommand\FigsGridHspace{-0.22cm}
\newcommand\FigsGridHspacee{0.22cm}

\newcommand\MID{6}
\newcommand\MIDbis{400}

\newcommand\NYUv{5}
\newcommand\NYUvbis{103}

\newcommand\DIMLv{326}
\newcommand\DIMLvbis{4856}

\begin{figure*}[t]
\vspace{-0.2cm}
  \centering
  \tiny	
% \begin{center}
% 	\hspace{1.5cm}\includegraphics[width=0.25\linewidth]{diffs/colorbar}\vspace{-.5cm}
% \end{center}

%0.141 0.106
\subfloat[{Guide.}]{
    \begin{minipage}[b]{\FigsGridWidth\linewidth} 
        
        \includegraphics[width=\linewidth]{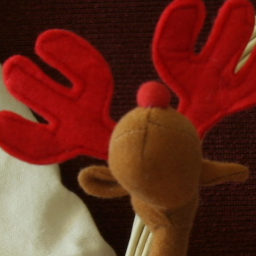}\vspace{\FigsGridVspace} 
        \includegraphics[width=\linewidth]{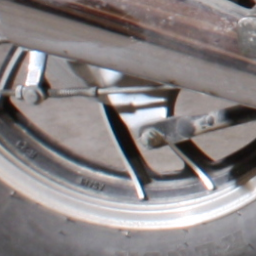}\vspace{\FigsGridVspace}\vspace{\FigsGridVspacee}
    
        \includegraphics[width=\linewidth]{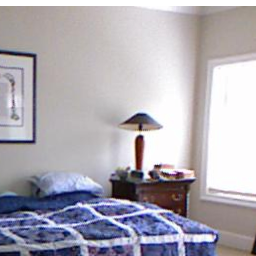}\vspace{\FigsGridVspace}
        \includegraphics[width=\linewidth]{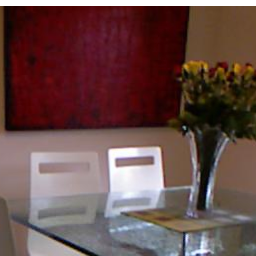}\vspace{\FigsGridVspace}\vspace{\FigsGridVspacee}
    
    \includegraphics[width=\linewidth]{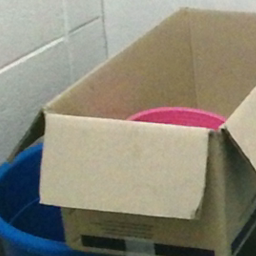}\vspace{\FigsGridVspace}
    \includegraphics[width=\linewidth]{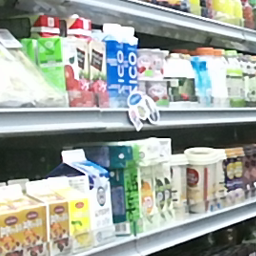}\vspace{\FigsGridVspace}
    \end{minipage}
}\hspace{\FigsGridHspace}
\subfloat[{Source}.]{

    \begin{minipage}[b]{\FigsGridWidth\linewidth} 
    \includegraphics[width=\linewidth]{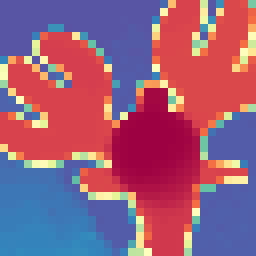}\vspace{\FigsGridVspace} 
    \includegraphics[width=\linewidth]{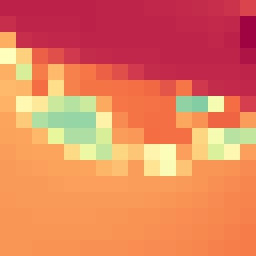}\vspace{\FigsGridVspace}\vspace{\FigsGridVspacee}
    
    \includegraphics[width=\linewidth]{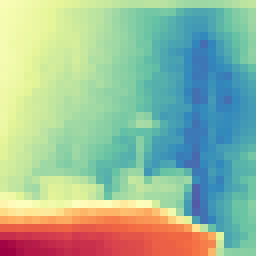}\vspace{\FigsGridVspace}
    \includegraphics[width=\linewidth]{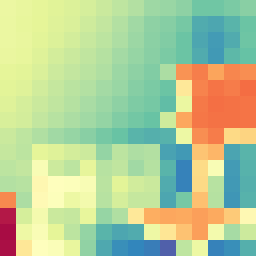}\vspace{\FigsGridVspace}\vspace{\FigsGridVspacee}
    
    \includegraphics[width=\linewidth]{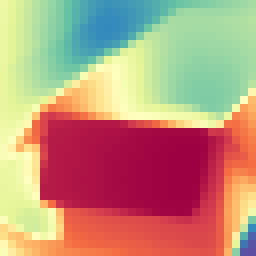}\vspace{\FigsGridVspace}
    \includegraphics[width=\linewidth]{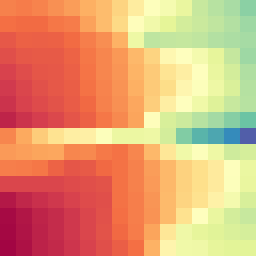}\vspace{\FigsGridVspace}
    \end{minipage}

}\hspace{\FigsGridHspace}
\subfloat[{GT}.]{
    \begin{minipage}[b]{\FigsGridWidth\linewidth} 
    \includegraphics[width=\linewidth]{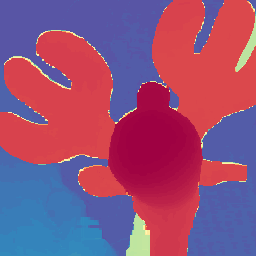}\vspace{\FigsGridVspace}
    \includegraphics[width=\linewidth]{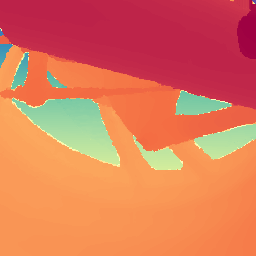}\vspace{\FigsGridVspace}\vspace{\FigsGridVspacee}
    
    \includegraphics[width=\linewidth]{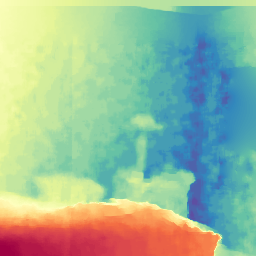}\vspace{\FigsGridVspace}
    \includegraphics[width=\linewidth]{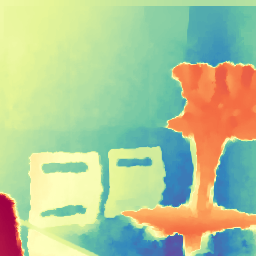}\vspace{\FigsGridVspace}\vspace{\FigsGridVspacee}
    
    \includegraphics[width=\linewidth]{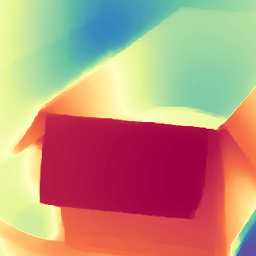}\vspace{\FigsGridVspace}
    \includegraphics[width=\linewidth]{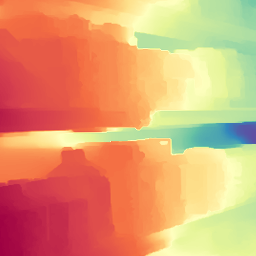}\vspace{\FigsGridVspace}
    \end{minipage}
}\hspace{\FigsGridHspace}\hspace{\FigsGridHspacee}
\subfloat[{SD}.]{
    \begin{minipage}[b]{\FigsGridWidth\linewidth} 
    \includegraphics[width=\linewidth]{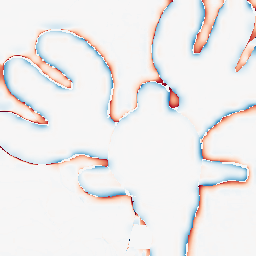}\vspace{\FigsGridVspace}
    \includegraphics[width=\linewidth]{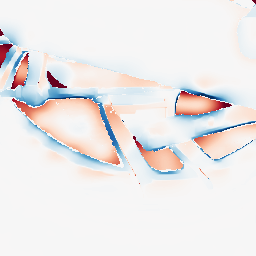}\vspace{\FigsGridVspace}\vspace{\FigsGridVspacee}
    
    \includegraphics[width=\linewidth]{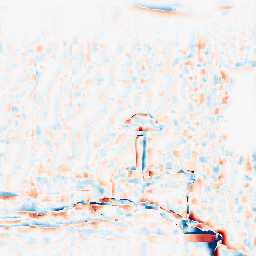}\vspace{\FigsGridVspace}
    \includegraphics[width=\linewidth]{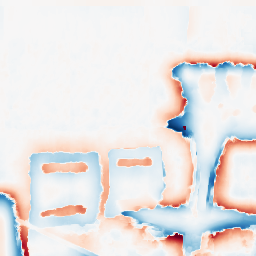}\vspace{\FigsGridVspace}\vspace{\FigsGridVspacee}
    
    \includegraphics[width=\linewidth]{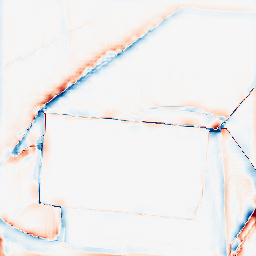}\vspace{\FigsGridVspace}
    \includegraphics[width=\linewidth]{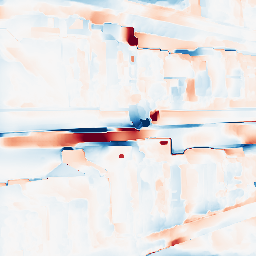}\vspace{\FigsGridVspace}
    \end{minipage}
}\hspace{\FigsGridHspace}
\subfloat[{Pixtr}.]{
    \begin{minipage}[b]{\FigsGridWidth\linewidth} 
    \includegraphics[width=\linewidth]{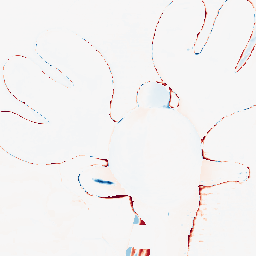}\vspace{\FigsGridVspace}
    \includegraphics[width=\linewidth]{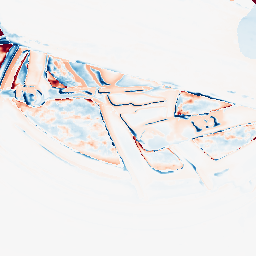}\vspace{\FigsGridVspace}\vspace{\FigsGridVspacee}  
    
    \includegraphics[width=\linewidth]{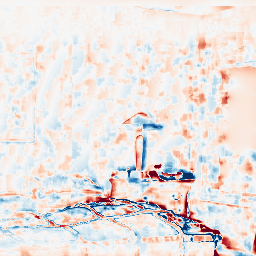}\vspace{\FigsGridVspace}
    \includegraphics[width=\linewidth]{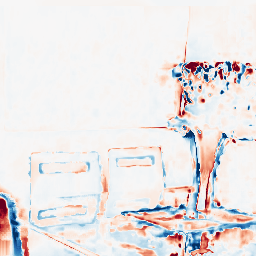}\vspace{\FigsGridVspace}\vspace{\FigsGridVspacee}
    
    \includegraphics[width=\linewidth]{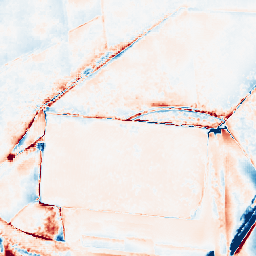}\vspace{\FigsGridVspace}
    \includegraphics[width=\linewidth]{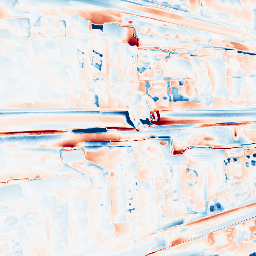}\vspace{\FigsGridVspace}
    \end{minipage}
}\hspace{\FigsGridHspace}
\subfloat[{MSG}.]{
    \begin{minipage}[b]{\FigsGridWidth\linewidth} 
    \includegraphics[width=\linewidth]{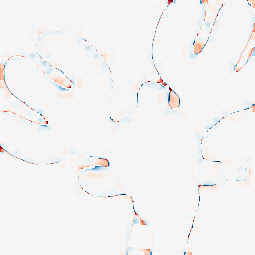}\vspace{\FigsGridVspace}
    \includegraphics[width=\linewidth]{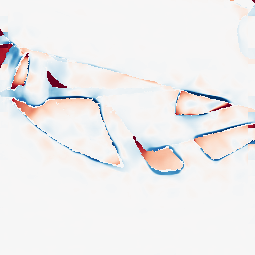}\vspace{\FigsGridVspace}\vspace{\FigsGridVspacee}
    
    \includegraphics[width=\linewidth]{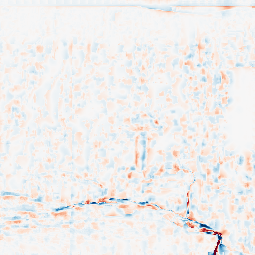}\vspace{\FigsGridVspace}
    \includegraphics[width=\linewidth]{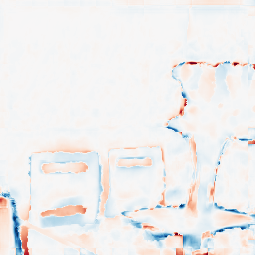}\vspace{\FigsGridVspace}\vspace{\FigsGridVspacee}
    
    \includegraphics[width=\linewidth]{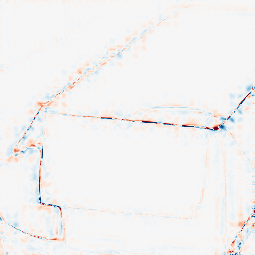}\vspace{\FigsGridVspace}
    \includegraphics[width=\linewidth]{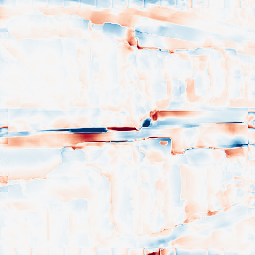}\vspace{\FigsGridVspace}
    \end{minipage}
}\hspace{\FigsGridHspace}
\subfloat[{FDKN}.]{
    \begin{minipage}[b]{\FigsGridWidth\linewidth} 
    \includegraphics[width=\linewidth]{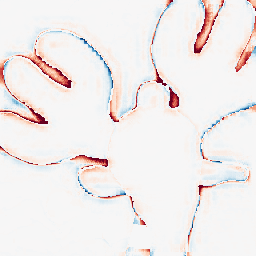}\vspace{\FigsGridVspace}
    \includegraphics[width=\linewidth]{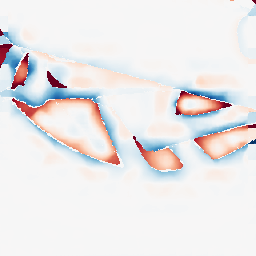}\vspace{\FigsGridVspace}\vspace{\FigsGridVspacee}
    
    \includegraphics[width=\linewidth]{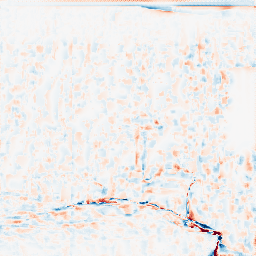}\vspace{\FigsGridVspace}
    \includegraphics[width=\linewidth]{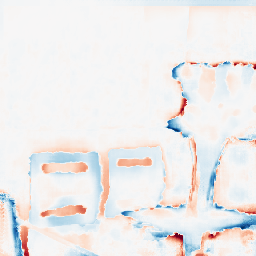}\vspace{\FigsGridVspace}\vspace{\FigsGridVspacee}
    
    \includegraphics[width=\linewidth]{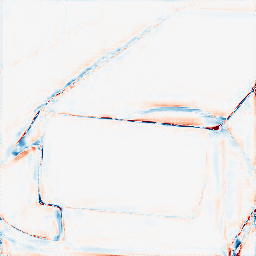}\vspace{\FigsGridVspace}
    \includegraphics[width=\linewidth]{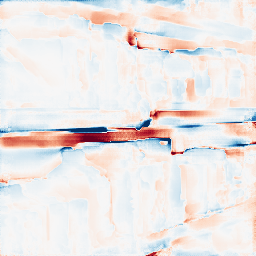}\vspace{\FigsGridVspace}
    \end{minipage}
}\hspace{\FigsGridHspace}
\subfloat[{PMBA}.]{
    \begin{minipage}[b]{\FigsGridWidth\linewidth} 
    \includegraphics[width=\linewidth]{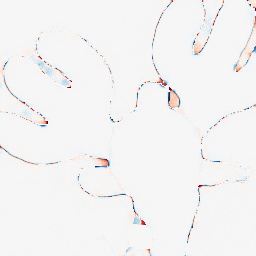}\vspace{\FigsGridVspace}
    \includegraphics[width=\linewidth]{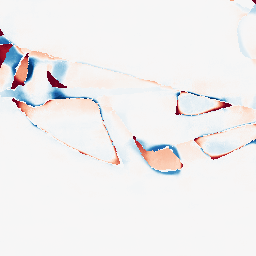}\vspace{\FigsGridVspace}\vspace{\FigsGridVspacee}
    
    \includegraphics[width=\linewidth]{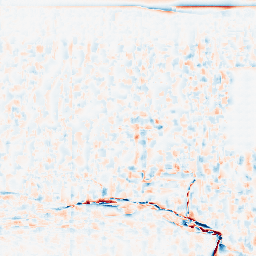}\vspace{\FigsGridVspace}
    \includegraphics[width=\linewidth]{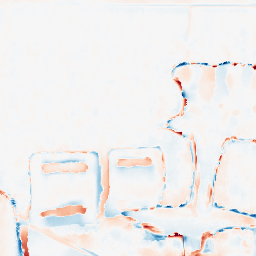}\vspace{\FigsGridVspace}\vspace{\FigsGridVspacee}
    
    \includegraphics[width=\linewidth]{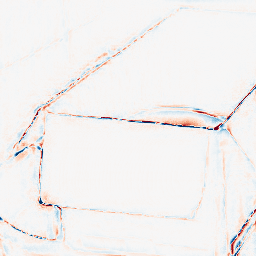}\vspace{\FigsGridVspace}
    \includegraphics[width=\linewidth]{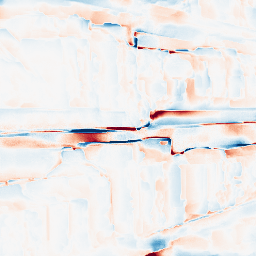}\vspace{\FigsGridVspace}
    \end{minipage}
}\hspace{\FigsGridHspace}
\subfloat[{FDSR}.]{
    \begin{minipage}[b]{\FigsGridWidth\linewidth} 
    \includegraphics[width=\linewidth]{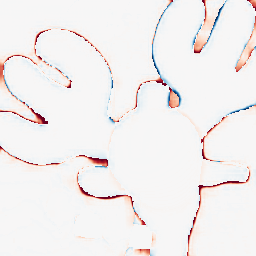}\vspace{\FigsGridVspace}
    \includegraphics[width=\linewidth]{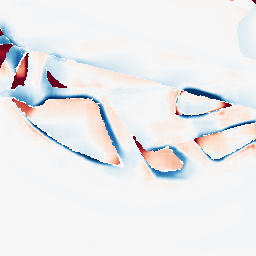}\vspace{\FigsGridVspace}\vspace{\FigsGridVspacee}
    
    \includegraphics[width=\linewidth]{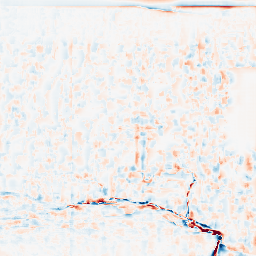}\vspace{\FigsGridVspace}
    \includegraphics[width=\linewidth]{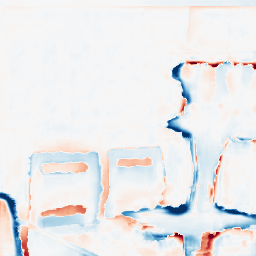}\vspace{\FigsGridVspace}\vspace{\FigsGridVspacee}
    
    \includegraphics[width=\linewidth]{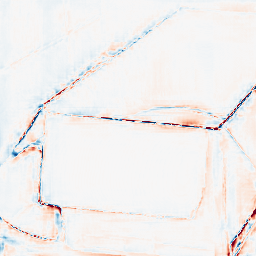}\vspace{\FigsGridVspace}
    \includegraphics[width=\linewidth]{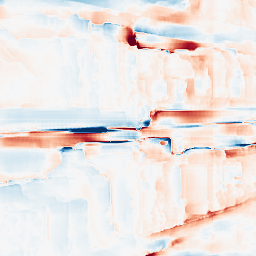}\vspace{\FigsGridVspace}
    \end{minipage}
}\hspace{\FigsGridHspace}\hspace{\FigsGridHspacee}
\subfloat[{Ours}.]{
    \begin{minipage}[b]{\FigsGridWidth\linewidth} 
    \includegraphics[width=\linewidth]{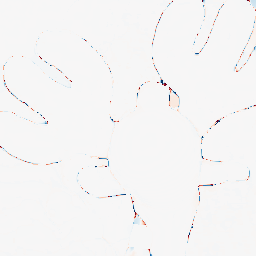}\vspace{\FigsGridVspace}
    \includegraphics[width=\linewidth]{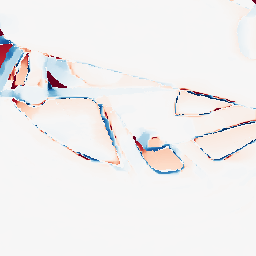}\vspace{\FigsGridVspace}\vspace{\FigsGridVspacee}
    
    \includegraphics[width=\linewidth]{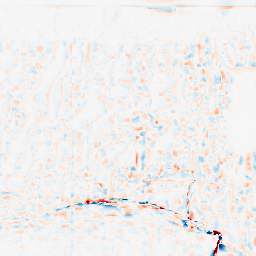}\vspace{\FigsGridVspace}
    \includegraphics[width=\linewidth]{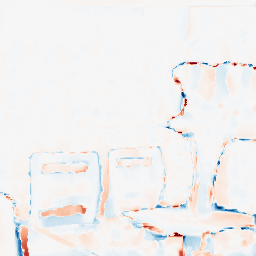}\vspace{\FigsGridVspace}\vspace{\FigsGridVspacee}
    
    \includegraphics[width=\linewidth]{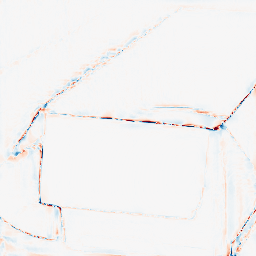}\vspace{\FigsGridVspace}
    \includegraphics[width=\linewidth]{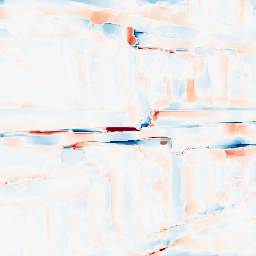}\vspace{\FigsGridVspace}
    \end{minipage}
}\hspace{\FigsGridHspace}
\subfloat[{Ours Pred}.]{
    \begin{minipage}[b]{\FigsGridWidth\linewidth} 
    \includegraphics[width=\linewidth]{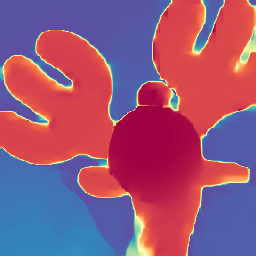}\vspace{\FigsGridVspace}
    \includegraphics[width=\linewidth]{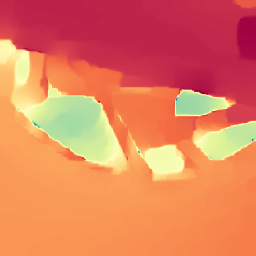}\vspace{\FigsGridVspace}\vspace{\FigsGridVspacee}
    
    \includegraphics[width=\linewidth]{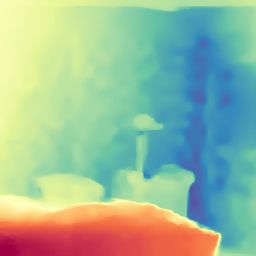}\vspace{\FigsGridVspace}
    \includegraphics[width=\linewidth]{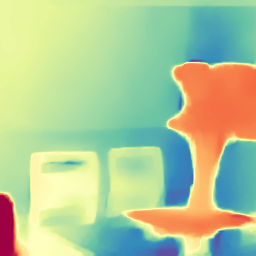}\vspace{\FigsGridVspace}\vspace{\FigsGridVspacee}
    
    \includegraphics[width=\linewidth]{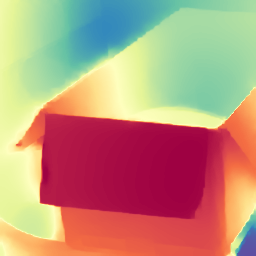}\vspace{\FigsGridVspace}
    \includegraphics[width=\linewidth]{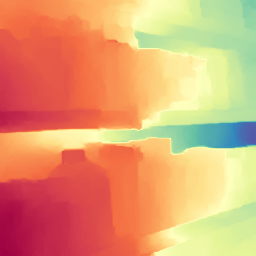}\vspace{\FigsGridVspace}
    \end{minipage}
}
\caption{Qualitative comparison of upsampled depth maps. From top to bottom each group of two rows shows the error of upsampled images, defined as the difference between the prediction and the ground truth, on the Middlebury \cite{Scharstein2007,Scharstein2001,Scharstein2014,Scharstein2003,Hirschmuller2007}, NYUv2 \cite{Silberman2012} and DIML \cite{DIML1,DIML2,DIML3,DIML4} datasets, respectively; alternating between upsampling factors $\times8$ and $\times16$ for each dataset. From left to right, the first group of columns are (a) Guide, (b) Source and (c) Ground Truth; the second group includes selected methods from our quantitative evaluation, (d) SD filter \cite{Ham2018}, (e) Pixtransform \cite{deLutio2019}, (f) MSG-Net \cite{Tak-Wai2016}, (g) FDKN \cite{Kim2021}, (h) PMBANet \cite{Ye2020} and (i) FDSR \cite{he2021}; the last two columns represent (j) the error for the prediction of our model and (k) the prediction itself.}
\label{fig:depth_upsampling}
\vspace{-0.3cm}
\end{figure*}

\subsection{Feature Extractor Comparison}
We go on to investigate the performance of our method with different feature extractors. Table~\ref{tab:ablation} compares the errors on the NYUv2 $\times8$ upsampling task with different backbones used to extract the features for the graph regularisation layer, as described in Section~\ref{sec:graph_reg}. We have tested several well-known backbones, always initialising them with weights pretrained on ImageNet \cite{deng2009imagenet}. In addition to these generic backbones, we also evaluate feature maps extracted from the FDSR network that was specifically designed for guided depth super-resolution.
Finally, to explore the boundaries of feature extraction, we employ a variant of the \emph{Dense Prediction Transformer} \cite{Ranftl2020, Ranftl2021}, which extracts multi-scale features for dense prediction tasks (like mono-depth or semantic segmentation) from a Vision Transformer (ViT). We adapted the model to account for the (appropriately re-sampled) source image at each feature level and call it the \emph{Guided Dense Prediction Transformer} (GDPT).
\begin{table}[b]
	\begin{center}
	\resizebox{\linewidth}{!}{
		\begin{tabular}{lccc}
			Feature extractor & \# Params & MSE ($\text{cm}^2$) & MAE ($\text{cm}$)\\
			\hline\hline
			Colour & 2 & 68.4 & 2.30 \\
		    UEfficientNet-B2 \cite{tan2019efficientnet} & 10M & 24.9 & 1.63 \\
			UResNet-18 \cite{he2016deep} & 14M & 21.7 & 1.52 \\
			UResNet-50 \cite{he2016deep} & 32M & \textbf{19.6} & \textbf{1.42} \\
			\hline\hline
			FDSR \cite{he2021} & 0.6M & 30.4 & 1.75 \\
			GDPT \cite{Ranftl2020, Ranftl2021} & 127M & 22.3 & 1.54 \\
            \hline\hline
		\end{tabular}}
	    \caption{Performance comparison of various feature extractors for the $\times8$ upsampling task on NYUv2.
	    }
		\label{tab:ablation}
	\end{center}
\end{table}
The results show that our method is fairly insensitive to the choice of architecture, across a range of capacities (respectively, parameter counts). This seems to indicate that the graph-based regularisation, although clearly benefiting from high-level features, bounds the expressive power of the feature extractor. We speculate that this is due to the fact that the graph cannot represent long-range patterns in the output image, but only enforces local smoothness where needed, thus limiting the amount of information that can be usefully transferred to the predicted target.
However, we do not recommend very low capacity backbones: when using FDSR, the performance is better than the original FDSR model, but clearly lower than with higher capacity models. 
Nevertheless, independent of the backbone used, our approach achieves the lowest MAE among all evaluated methods; except when using the raw \emph{Colour} as features, i.e., low-level image contrast is not sufficient as a regulariser and using a learned feature extractor is essential.
\section{Discussion}
The graph regularisation layer effectively acts as a bottleneck on the amount of information that can be carried from the guide to the target -- the regularisation is not able to create arbitrary patterns in the target image. This can be seen as a limitation, but also a desirable property, as it increases model robustness. 
One drawback of our method \wrt to most of the conventional deep forward architectures is the inference time, which is the price to pay for an online optimisation that guarantees rigorous fidelity to the source. A forward pass on a single $256^2$ pixel patch, for upsampling factor $\times 8$, takes on average $111\,$ms on an NVIDIA GeForce RTX 2080 Ti. This number varies depending on the complexity of the image, and the upsampling factor.

\section{Conclusion}
\label{sec:conclusion}

We have presented a novel formulation for guided super-resolution based on a learnable graph regulariser. The method employs a deep feature extractor that takes a \textit{guide} and a \textit{source} as input, and infers an affinity graph over adjacent pixels in the \textit{target} image. The learned graph serves as a regulariser in the upsampling of the source, implemented as a differentiable optimisation layer. This explicit optimisation within the architecture guarantees rigorous fidelity of the high-resolution target to the low-resolution source. 
Our proposed method combines desirable properties from both, conventional and deep learning based methods: the optimisation layer guarantees that the fidelity \wrt the source image is satisfied, even in case of domain shifts in the test set, while the deep feature extractor enables the learned affinity graph to encapsulate valuable information extracted from a large context. The experimental evaluation confirms that our graph regulariser is effective for signals that exhibit a piecewise smooth structure, such as depth maps. 

%%%%%%%%% REFERENCES
{\small
\bibliographystyle{ieee_fullname}
\bibliography{mybib}

\begin{thebibliography}{10}\itemsep=-1pt

\bibitem{Barron2015}
Jonathan~T. Barron, Andrew Adams, YiChang Shih, and Carlos Hern{\'a}ndez.
\newblock Fast bilateral-space stereo for synthetic defocus.
\newblock In {\em CVPR}, 2015.

\bibitem{Barron2016}
Jonathan~T. Barron and Ben Poole.
\newblock The fast bilateral solver.
\newblock In {\em ECCV}, 2016.

\bibitem{Chan2008}
Derek Chan, Hylke Buisman, Christian Theobalt, and Sebastian Thrun.
\newblock A noise-aware filter for real-time depth upsampling.
\newblock In {\em {Workshop on Multi-camera and Multi-modal Sensor Fusion
  Algorithms and Applications - M2SFA2}}, 2008.

\bibitem{DIML4}
{Cho, Jaehoon and Min, Dongbo and Kim, Youngjung and Sohn, Kwanghoon}.
\newblock Deep monocular depth estimation leveraging a large-scale outdoor
  stereo dataset.
\newblock {\em Expert Systems with Applications}, 2021.

\bibitem{deLutio2019}
Riccardo de Lutio, Stefano D'Aronco, Jan~D. Wegner, and Konrad Schindler.
\newblock Guided super-resolution as pixel-to-pixel transformation.
\newblock In {\em ICCV}, 2019.

\bibitem{deng2009imagenet}
Jia Deng, Wei Dong, Richard Socher, Li-Jia Li, Kai Li, and Li Fei-Fei.
\newblock Imagenet: A large-scale hierarchical image database.
\newblock In {\em CVPR}. Ieee, 2009.

\bibitem{Diebel2006}
James Diebel and Sebastian Thrun.
\newblock An application of markov random fields to range sensing.
\newblock In {\em NIPS}, 2006.

\bibitem{dong2016color}
Weisheng Dong, Guangming Shi, Xin Li, Kefan Peng, Jinjian Wu, and Zhenhua Guo.
\newblock Color-guided depth recovery via joint local structural and nonlocal
  low-rank regularization.
\newblock {\em IEEE Transactions on Multimedia}, 2016.

\bibitem{zheng2015}
Zheng et al.
\newblock Conditional random fields as recurrent neural networks.
\newblock In {\em ICLR}, 2015.

\bibitem{Ferstl2013}
David Ferstl, Christian Reinbacher, Rene Ranftl, Matthias R\"{u}ther, and Horst
  Bischof.
\newblock Image guided depth upsampling using anisotropic total generalized
  variation.
\newblock In {\em ICCV}, 2013.

\bibitem{Guo2019}
Chunle Guo, Chongyi Li, Jichang Guo, Runmin Cong, Huazhu Fu, and Ping Han.
\newblock Hierarchical features driven residual learning for depth map
  super-resolution.
\newblock {\em TIP}, 2019.

\bibitem{Ham2018}
Bumsub Ham, Minsu Cho, and Jean Ponce.
\newblock Robust guided image filtering using nonconvex potentials.
\newblock {\em TPAMI}, 2018.

\bibitem{He2013}
Kaiming He, Jian Sun, and Xiaoou Tang.
\newblock Guided image filtering.
\newblock {\em TPAMI}, 2013.

\bibitem{he2016deep}
Kaiming He, Xiangyu Zhang, Shaoqing Ren, and Jian Sun.
\newblock Deep residual learning for image recognition.
\newblock In {\em CVPR}, 2016.

\bibitem{he2021}
Lingzhi He, Hongguang Zhu, Feng Li, Huihui Bai, Runmin Cong, Chunjie Zhang,
  Chunyu Lin, Meiqin Liu, and Yao Zhao.
\newblock Towards fast and accurate real-world depth super-resolution:
  Benchmark dataset and baseline.
\newblock In {\em CVPR}, 2021.

\bibitem{Hirschmuller2007}
Heiko Hirschm{\"u}ller and Daniel Scharstein.
\newblock Evaluation of cost functions for stereo matching.
\newblock In {\em CVPR}, 2007.

\bibitem{Tak-Wai2016}
Tak-Wai Hui, Chen~Change Loy, and Xiaoou Tang.
\newblock Depth map super-resolution by deep multi-scale guidance.
\newblock In {\em ECCV}, 2016.

\bibitem{Kim2021}
Beomjun Kim, Jean Ponce, and Bumsub Ham.
\newblock Deformable kernel networks for joint image filtering.
\newblock {\em IJCV}, 2021.

\bibitem{kim2016accurate}
Jiwon Kim, Jung~Kwon Lee, and Kyoung~Mu Lee.
\newblock Accurate image super-resolution using very deep convolutional
  networks.
\newblock In {\em CVPR}, 2016.

\bibitem{DIML2}
Sunok Kim, Dongbo Min, Bumsub Ham, Seungryong Kim, and Kwanghoon Sohn.
\newblock Deep stereo confidence prediction for depth estimation.
\newblock In {\em ICIP}, 2017.

\bibitem{DIML1}
Youngjung Kim, Bumsub Ham, Changjae Oh, and Kwanghoon Sohn.
\newblock Structure selective depth superresolution for rgb-d cameras.
\newblock {\em TIP}, 2016.

\bibitem{DIML3}
Youngjung Kim, Hyungjoo Jung, Dongbo Min, and Kwanghoon Sohn.
\newblock Deep monocular depth estimation via integration of global and local
  predictions.
\newblock {\em TIP}, 2018.

\bibitem{Kingma2014}
Diederik~P. Kingma and Jimmy Ba.
\newblock Adam: A method for stochastic optimization.
\newblock In {\em ICLR}, 2015.

\bibitem{Kopf2007}
Johannes Kopf, Michael~F. Cohen, Dani Lischinski, and Matt Uyttendaele.
\newblock Joint bilateral upsampling.
\newblock {\em ToG}, 2007.

\bibitem{Kwon2015}
HyeokHyen Kwon, Yu-Wing Tai, and Stephen Lin.
\newblock Data-driven depth map refinement via multi-scale sparse
  representation.
\newblock In {\em CVPR}, 2015.

\bibitem{Lanaras2018}
Charis Lanaras, Jos{\'e} Bioucas-Dias, Silvano Galliani, Emmanuel Baltsavias,
  and Konrad Schindler.
\newblock Super-resolution of {Sentinel-2} images: Learning a globally
  applicable deep neural network.
\newblock {\em ISPRS Journal of Photogrammetry and Remote Sensing}, 2018.

\bibitem{Li2016}
Yijun Li, Jia-Bin Huang, Narendra Ahuja, and Ming-Hsuan Yang.
\newblock Deep joint image filtering.
\newblock In {\em ECCV}, 2016.

\bibitem{Li2019}
Yijun Li, Jia-Bin Huang, Narendra Ahuja, and Ming-Hsuan Yang.
\newblock Joint image filtering with deep convolutional networks.
\newblock {\em TPAMI}, 2019.

\bibitem{Li2012}
Yanjie Li, Tianfan Xue, Lifeng Sun, and Jianzhuang Liu.
\newblock Joint example-based depth map super-resolution.
\newblock In {\em ICME}, 2012.

\bibitem{Liu2013}
Ming-Yu Liu, Oncel Tuzel, and Yuichi Taguchi.
\newblock Joint geodesic upsampling of depth images.
\newblock In {\em CVPR}, 2013.

\bibitem{Silberman2012}
Pushmeet~Kohli Nathan~Silberman, Derek~Hoiem and Rob Fergus.
\newblock Indoor segmentation and support inference from rgbd images.
\newblock In {\em ECCV}, 2012.

\bibitem{numerical_opt}
Jorge Nocedal and Stephen~J. Wright.
\newblock {\em Numerical Optimization}.
\newblock Springer, 2006.

\bibitem{Okuta2017}
Ryosuke Okuta, Yuya Unno, Daisuke Nishino, Shohei Hido, and Crissman Loomis.
\newblock Cupy: A numpy-compatible library for nvidia gpu calculations.
\newblock In {\em Proceedings of Workshop on Machine Learning Systems
  (LearningSys) in The Thirty-first Annual Conference on Neural Information
  Processing Systems (NIPS)}, 2017.

\bibitem{Pan2019}
Jinshan Pan, Jiangxin Dong, Jimmy~S. Ren, Liang Lin, Jinhui Tang, and
  Ming-Hsuan Yang.
\newblock Spatially variant linear representation models for joint filtering.
\newblock In {\em CVPR}, 2019.

\bibitem{Park2011}
Jaesik Park, Hyeongwoo Kim, Yu-Wing Tai, Michael~S. Brown, and Inseo Kweon.
\newblock High quality depth map upsampling for {3D-TOF} cameras.
\newblock In {\em ICCV}, 2011.

\bibitem{PARK201950}
Min-Gyu Park and Kuk-Jin Yoon.
\newblock As-planar-as-possible depth map estimation.
\newblock {\em Computer Vision and Image Understanding}, 2019.

\bibitem{Paszke2019}
Adam Paszke, Sam Gross, Francisco Massa, Adam Lerer, James Bradbury, Gregory
  Chanan, Trevor Killeen, Zeming Lin, Natalia Gimelshein, Luca Antiga, Alban
  Desmaison, Andreas Kopf, Edward Yang, Zachary DeVito, Martin Raison, Alykhan
  Tejani, Sasank Chilamkurthy, Benoit Steiner, Lu Fang, Junjie Bai, and Soumith
  Chintala.
\newblock Pytorch: An imperative style, high-performance deep learning library.
\newblock In {\em NIPS}, 2019.

\bibitem{Ranftl2021}
Ren\'{e} Ranftl, Alexey Bochkovskiy, and Vladlen Koltun.
\newblock Vision transformers for dense prediction.
\newblock {\em ArXiv preprint}, 2021.

\bibitem{Ranftl2020}
Ren\'{e} Ranftl, Katrin Lasinger, David Hafner, Konrad Schindler, and Vladlen
  Koltun.
\newblock Towards robust monocular depth estimation: Mixing datasets for
  zero-shot cross-dataset transfer.
\newblock {\em TPAMI}, 2020.

\bibitem{Riegler2016}
Gernot Riegler, David Ferstl, Matthias R\"{u}ther, and Horst Bischof.
\newblock A deep primal-dual network for guided depth super-resolution.
\newblock In {\em BMVC}, 2016.

\bibitem{ronneberger2015u}
Olaf Ronneberger, Philipp Fischer, and Thomas Brox.
\newblock U-net: Convolutional networks for biomedical image segmentation.
\newblock In {\em International Conference on Medical Image Computing and
  Computer-Assisted Intervention}, 2015.

\bibitem{Rossi2020}
Mattia Rossi, Mireille {El Gheche}, Andreas Kuhn, and Pascal Frossard.
\newblock Joint graph-based depth refinement and normal estimation.
\newblock In {\em CVPR}, 2020.

\bibitem{safin2021unpaired}
Aleksandr Safin, Maxim Kan, Nikita Drobyshev, Oleg Voynov, Alexey Artemov,
  Alexander Filippov, Denis Zorin, and Evgeny Burnaev.
\newblock Unpaired depth super-resolution in the wild, 2021.

\bibitem{Scharstein2014}
Daniel Scharstein, Heiko Hirschm{\"u}ller, York Kitajima, Greg Krathwohl, Nera
  Ne{\v{s}}i{\'{c}}, Xi Wang, and Porter Westling.
\newblock High-resolution stereo datasets with subpixel-accurate ground truth.
\newblock In {\em GCPR}, 2014.

\bibitem{Scharstein2007}
Daniel Scharstein and Chris Pal.
\newblock Learning conditional random fields for stereo.
\newblock In {\em CVPR}, 2007.

\bibitem{Scharstein2003}
Daniel Scharstein and Richard Szeliski.
\newblock High-accuracy stereo depth maps using structured light.
\newblock In {\em CVPR}, 2001.

\bibitem{Scharstein2001}
Daniel Scharstein, Richard Szeliski, and Ramin Zabih.
\newblock A taxonomy and evaluation of dense two-frame stereo correspondence
  algorithms.
\newblock {\em IJCV}, 2002.

\bibitem{Shacht_2021_CVPR}
Guy Shacht, Dov Danon, Sharon Fogel, and Daniel Cohen-Or.
\newblock Single pair cross-modality super resolution.
\newblock In {\em CVPR}, 2021.

\bibitem{Shelhamer2017}
Evan Shelhamer, Jonathan Long, and Trevor Darrell.
\newblock Fully convolutional networks for semantic segmentation.
\newblock {\em TPAMI}, 2017.

\bibitem{Song2020l}
Xibin Song, Yuchao Dai, Dingfu Zhou, Liu Liu, Wei Li, Hongdng Li, and Ruigang
  Yang.
\newblock Channel attention based iterative residual learning for depth map
  super-resolution.
\newblock In {\em CVPR}, 2020.

\bibitem{su2019}
Hang Su, Varun Jampani, Deqing Sun, Orazio Gallo, Erik Learned-Miller, and Jan
  Kautz.
\newblock Pixel-adaptive convolutional neural networks.
\newblock In {\em CVPR}, 2019.

\bibitem{Sun2021cvpr}
Baoli Sun, Xinchen Ye, Baopu Li, Haojie Li, Zhihui Wang, and Rui Xu.
\newblock Learning scene structure guidance via cross-task knowledge transfer
  for single depth super-resolution.
\newblock In {\em CVPR}, 2021.

\bibitem{tan2019efficientnet}
Mingxing Tan and Quoc Le.
\newblock Efficientnet: Rethinking model scaling for convolutional neural
  networks.
\newblock In {\em International Conference on Machine Learning}. PMLR, 2019.

\bibitem{dataset_bias}
Antonio Torralba and Alexei~A. Efros.
\newblock Unbiased look at dataset bias.
\newblock In {\em CVPR}, 2011.

\bibitem{Uezato2020}
Tatsumi Uezato, Danfeng Hong, Naoto Yokoya, and Wei He.
\newblock Guided deep decoder: Unsupervised image pair fusion.
\newblock In {\em ECCV}, 2020.

\bibitem{Ulyanov2018}
Dmitry Ulyanov, Andrea Vedaldi, and Victor Lempitsky.
\newblock Deep image prior.
\newblock In {\em CVPR}, 2018.

\bibitem{Wen2019}
Yang Wen, Bin Sheng, Ping Li, Weiyao Lin, and David~Dagan Feng.
\newblock Deep color guided coarse-to-fine convolutional network cascade for
  depth image super-resolution.
\newblock {\em TIP}, 2019.

\bibitem{Yang2012}
Jingyu Yang, Xinchen Ye, Kun Li, and Chunping Hou.
\newblock Depth recovery using an adaptive color-guided auto-regressive model.
\newblock In {\em ECCV}, 2012.

\bibitem{Yang2014}
Jingyu Yang, Xinchen Ye, Kun Li, Chunping Hou, and Yao Wang.
\newblock Color-guided depth recovery from {RGB-D} data using an adaptive
  autoregressive model.
\newblock {\em TIP}, 2014.

\bibitem{Yang2007}
Qingxiong Yang, Ruigang Yang, James Davis, and David Nister.
\newblock Spatial-depth super resolution for range images.
\newblock In {\em CVPR}, 2007.

\bibitem{Ye2020}
Xinchen Ye, Baoli Sun, Zhihui Wang, Jingyu Yang, Rui Xu, Haojie Li, and Baopu
  Li.
\newblock {PMBANet}: Progressive multi-branch aggregation network for scene
  depth super-resolution.
\newblock {\em TIP}, 2020.

\bibitem{zhang2018longitudinally}
Yongqin Zhang, Feng Shi, Jian Cheng, Li Wang, Pew-Thian Yap, and Dinggang Shen.
\newblock Longitudinally guided super-resolution of neonatal brain magnetic
  resonance images.
\newblock {\em IEEE Transactions on Cybernetics}, 2018.

\end{thebibliography}
}

\newpage
%%%%%%%%% TITLE - 
%\title{\vspace{-1.5cm}Supplementary Material\\Learning Graph Regularisation for Guided Super-Resolution\vspace{-1cm}}  

\twocolumn[\Large \centering \textbf{Supplementary Material} \\\textbf{Learning Graph Regularisation for Guided Super-Resolution} \vspace{1cm}]

\appendix

\renewcommand{\thefigure}{A\arabic{figure}}
\renewcommand{\thetable}{A\arabic{table}}
\setcounter{figure}{0}
\setcounter{table}{0}

%%%%%%%%% BODY TEXT - 

%-------------------------------------------------------------------------
\section{Hyperparameters}
We report the hyperparameter settings for the evaluation of our proposed method for RGB-guided depth map super-resolution. Hyperparameters are reported for each dataset and method.
The experiments were conducted on the three RGB-D datasets Middlebury \cite{Scharstein2007,Scharstein2001,Scharstein2014,Scharstein2003,Hirschmuller2007}, NYUv2 \cite{Silberman2012} and DIML \cite{DIML1,DIML2,DIML3,DIML4} with the following methods: Guided Filter (GF) \cite{He2013}, the Static/Dynamic filter (SD) \cite{Ham2018}, the Pixtransform \cite{deLutio2019}, the MSG-Net \cite{Tak-Wai2016}, the Deformable Kernel Network (DKN) and its fast version (FDKN) \cite{Kim2021}, the PMBANet \cite{Ye2020}, and the Fast Depth Super-Resolution (FDSR) \cite{he2021}.

We train all learned methods using the Adam optimiser with default parameters $\beta_1=0.9$, $\beta_2=0.999$, $\epsilon=10^{-8}$ and a batch size of 8. For the Middlebury dataset, we train for 2,500 epochs with an initial learning rate of $10^{-4}$ and reduce it by factor 0.9 every 100 epochs. For the NYUv2 dataset, we train for 250 epochs with the same initial learning rate that is reduced by a factor of 0.9 every 10 epochs. For the DIML dataset, the methods are trained for 150 epochs, again using an initial learning rate of $10^{-4}$ with a reduction by a factor of 0.9 every 6 epochs. For all learned methods we additionally tested the initial learning rate and learning rate schedule proposed by the respective authors (if available), and used the best configuration. For FDSR, we therefore deviated from our default settings and chose an initial learning rate of $5.0\cdot10^{-4}$, with a 0.5$\times$ reduction every 80,000 \textit{iterations}. For PMBANet, we found a 0.1$\times$ reduction every 100 (1000, 60) epochs for NYUv2 (Middlebury, DIML) to work best. Note that we did not conduct any hyperparameter search for our proposed method.

We also report the hyperparameters used for the non-learned approaches. We used the following values for the SD filter, which we found to be the best performing among the tested configurations: $\lambda=0.1$, $\sigma_g=60$ and $\sigma_u=30$.
For Pixtransform we used the hyperparameters suggested in the original manuscript.

%\input{hyperparams}

%-------------------------------------------------------------------------
\section{Qualitative Results}

In Figures \ref{sup_fig:depth_upsampling_middlebury}, \ref{sup_fig:depth_upsampling_nyu} and \ref{sup_fig:depth_upsampling_diml}, we provide additional examples for qualitative comparison between our method and selected methods from our quantitative evaluation; for the Middlebury, NYUv2 and DIML datasets respectively. We provide two additional examples for each upsampling factor and dataset combination. These results further confirm that the predictions obtained with our method compare favourably to existing methods.
In particular, we observe the magnitude of errors for our method to be smaller compared to the other approaches, especially along edges. For many examples, we can additionally see that our method facilitates smooth depth predictions in continuous areas, whereas other methods exhibit higher noise.
Visually, the differences seem small for an upsampling factor of $\times 4$, however with larger upsampling factors they become more apparent.

%------------------------------------------------------------------------
\section{Learning Graph Weights}

\begin{figure*}[h]
\centering

\subfloat{
    \begin{minipage}[b]{0.42\linewidth} 
    \includegraphics[width=\linewidth]{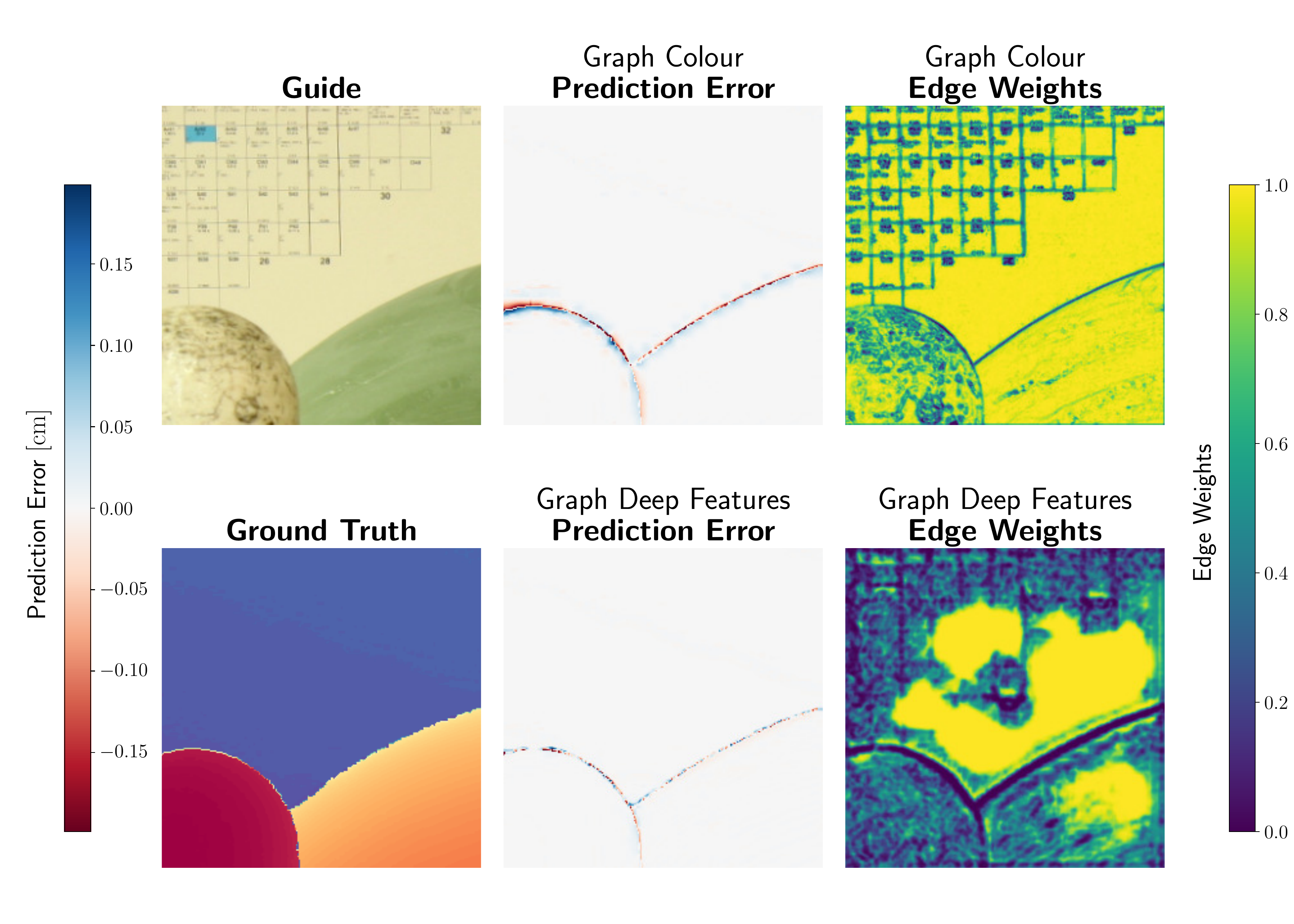}
    \includegraphics[width=\linewidth]{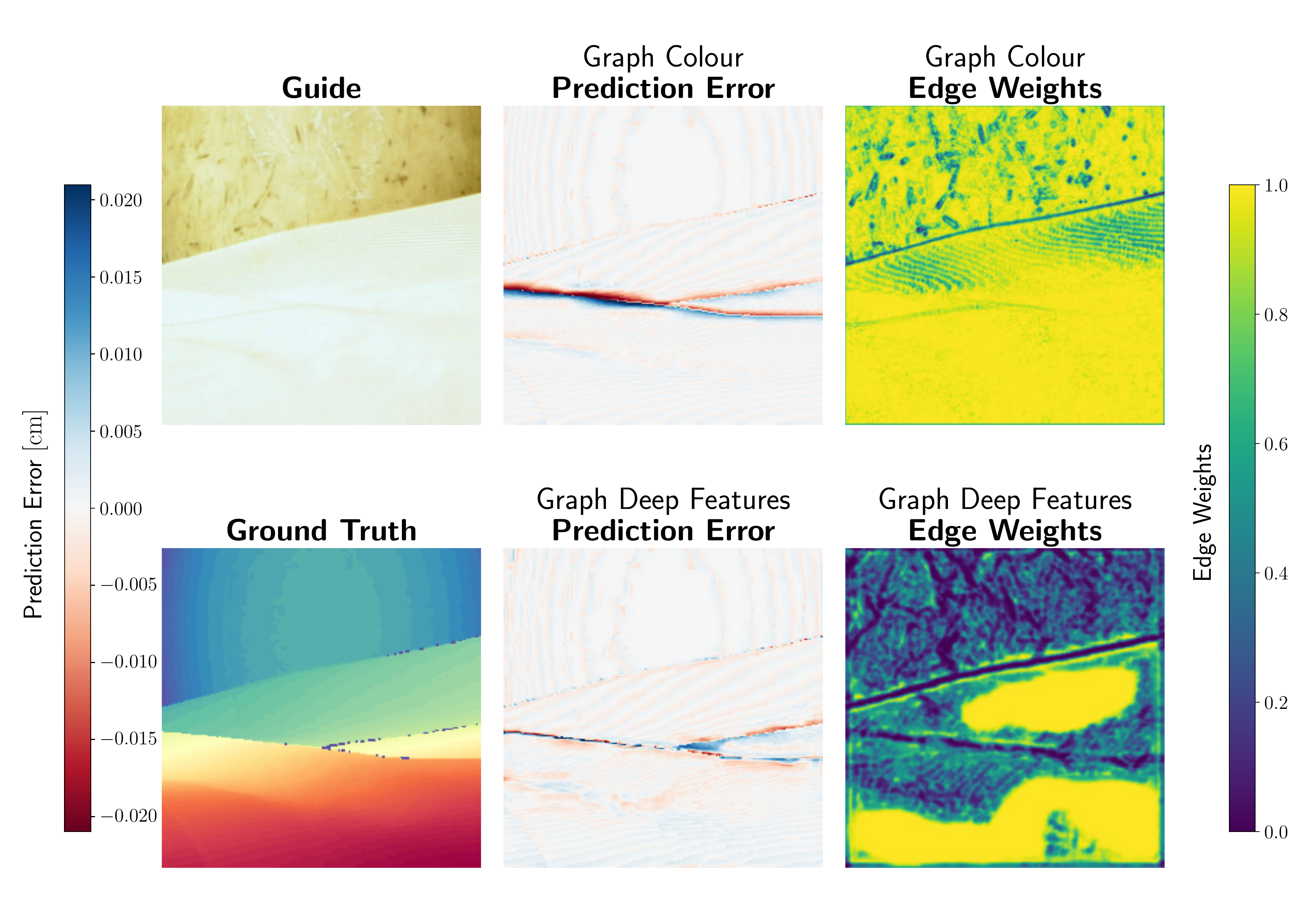}
    \end{minipage} }
    \subfloat{
    \begin{minipage}[b]{0.42\linewidth} 
    \includegraphics[width=\linewidth]{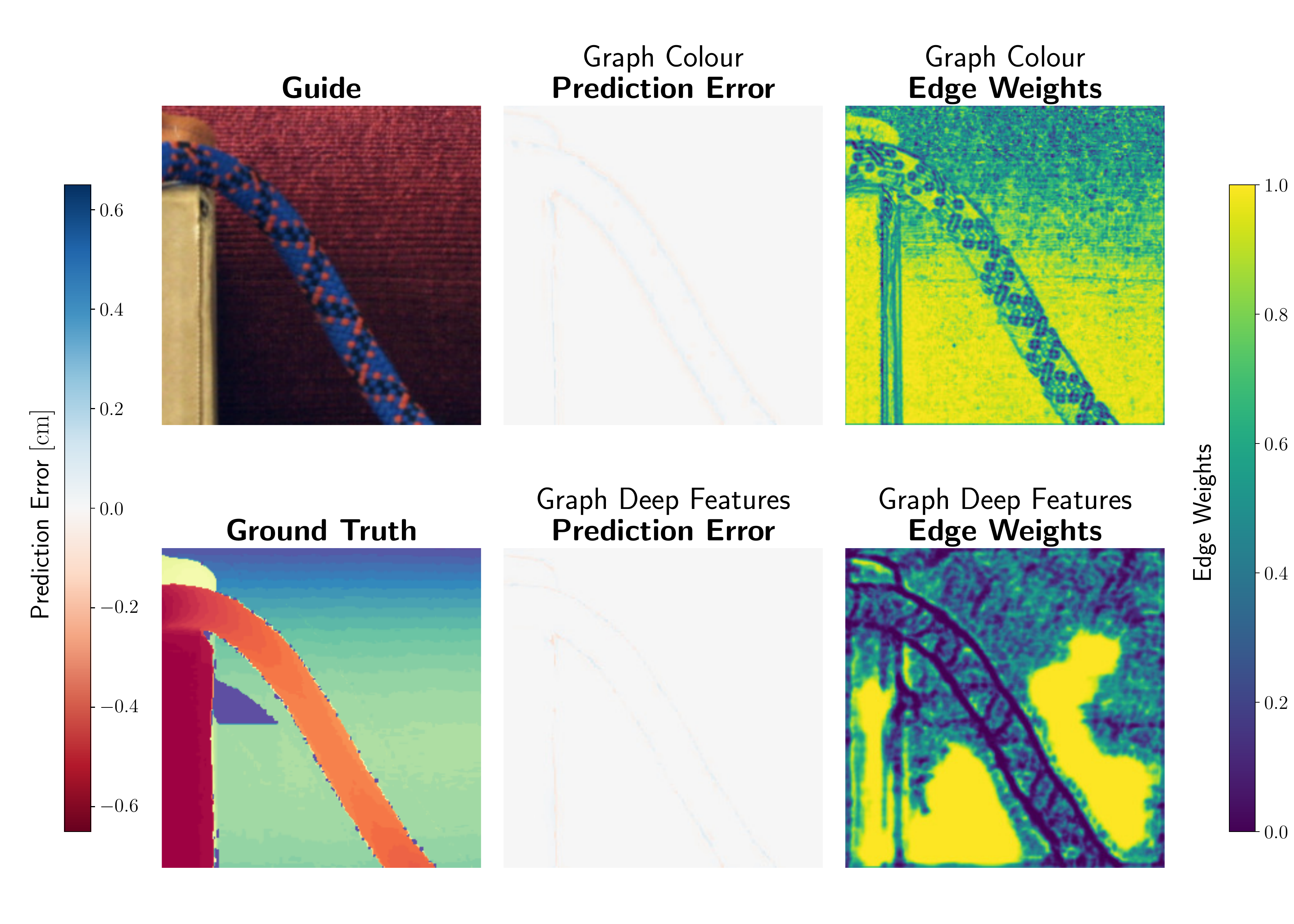}
    \includegraphics[width=\linewidth]{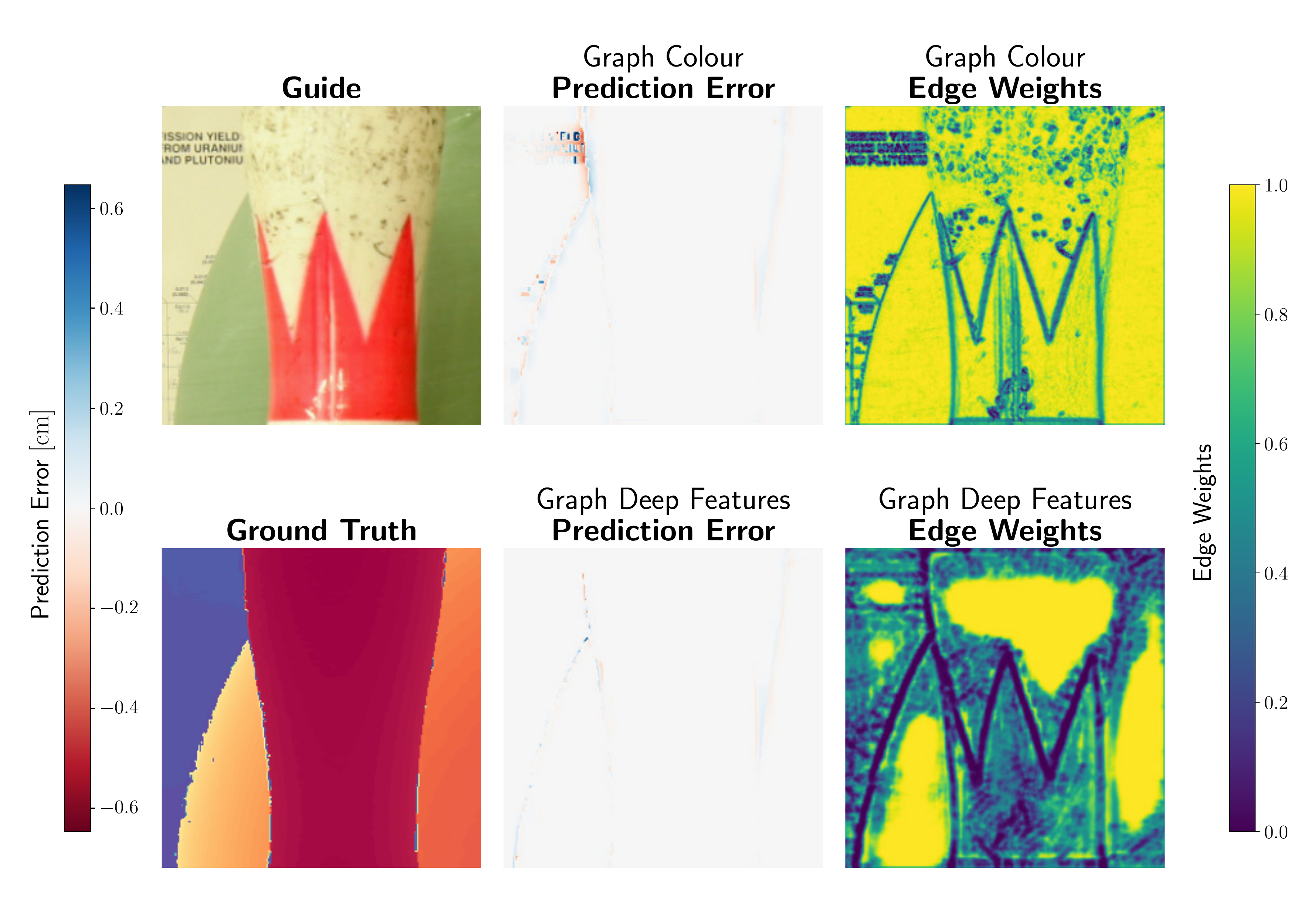}
    \end{minipage} }
    \caption{Additional examples of the importance of learned edge potentials. We visualise the total affinity of each pixel to its four neighbours when derived from raw colour (top) or from deep features (bottom). Examples are from the Middlebury test set.}
    \label{sup_fig:weights}
\end{figure*}

In Figure \ref{sup_fig:weights} we show additional examples of the difference between the total affinity of each pixel to its four neighbours when the graph is defined on colour features against the graph defined on learned features. These results further show that our model is able to encourage the optimiser to provide smooth predictions in areas without depth discontinuities. On the other hand, the model also shows low predicted weights in areas that should not be smoothed, thus providing crisper predictions.

%------------------------------------------------------------------------

\section{Forward Pass Timing}

We provide some time statistics for the forward pass of the proposed method in Table~\ref{tab:forward_time}.
For reference: FDKN \cite{Kim2021} takes about 10 ms for a forward pass independently of the scaling factor. The forward pass for the feature extractor of our proposed method also takes about 15 ms. Pixtransform~\cite{deLutio2019}, by far the slowest method among the compared ones, takes more than 120 seconds for a $256^2$ patch. Our Python implementation of the SD filter~\cite{Ham2018} takes few seconds (this also depends on the upsampling factor), although it is not directly comparable to our method as the SD filter implementation does not take advantage of GPU acceleration. 
The upsampling factor plays a major role for the runtime of our method, because larger upsampling factors lead to downsampling operators $\mD$ that are less sparse, increasing the time required to solve the linear system.
Note that although our method was implemented to leverage GPU parallelisation, there is still potential for further optimisations that could improve runtime performance.

\begin{table}[h]
\centering
\begin{tabular}{ c  ccc }
 & $\times 4$ & $\times 8$ & $\times 16$ \\ \hline \hline
 
forward time  & 79 (21) & 111 (35) & 305 (92) \\ \hline\hline

\end{tabular}%

\caption{Forward pass times of our proposed method. Numbers represent the mean time and (standard deviation) measured in milliseconds, computed over the NYUv2 test set on single patches of $256^2$ pixels, on an NVIDIA GeForce RTX 2080 Ti.}
\label{tab:forward_time}
\end{table}

\newcommand\FigGridWidth{0.09}
\newcommand\FigGridVspace{0.05cm}
\newcommand\FigGridVspacee{0.15cm}
\newcommand\FigGridHspace{-0.22cm}
\newcommand\FigGridHspacee{0.22cm}

\newcommand\Mid{296}
\newcommand\Mideight{233}
\newcommand\Midsixteen{17}

\newcommand\bisMid{378}
\newcommand\bisMideight{349}
\newcommand\bisMidsixteen{326}

\begin{sidewaysfigure*}[p]
\vspace{-0.2cm}
  \centering
  \tiny	
\subfloat[{Guide.}]{
    \begin{minipage}[b]{\FigGridWidth\linewidth} 
        \includegraphics[width=\linewidth]{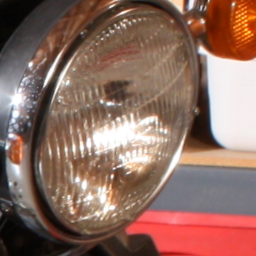}\vspace{\FigGridVspace}         
        \includegraphics[width=\linewidth]{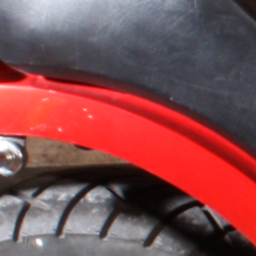}\vspace{\FigGridVspace}\vspace{\FigGridVspacee}        
        
        \includegraphics[width=\linewidth]{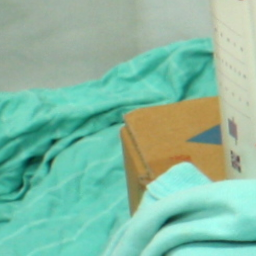}\vspace{\FigGridVspace} 
        \includegraphics[width=\linewidth]{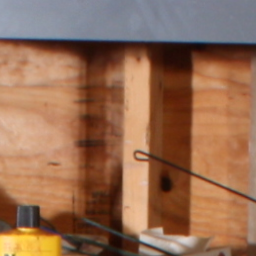}\vspace{\FigGridVspace}\vspace{\FigGridVspacee}   
   
        \includegraphics[width=\linewidth]{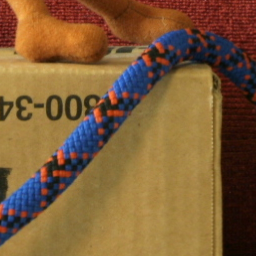}\vspace{\FigGridVspace}
        \includegraphics[width=\linewidth]{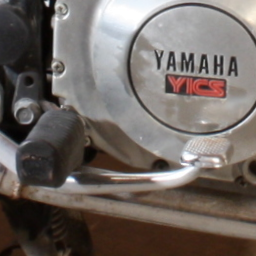}\vspace{\FigGridVspace}

    \end{minipage}
}\hspace{\FigGridHspace}
\subfloat[{Source}.]{

    \begin{minipage}[b]{\FigGridWidth\linewidth} 
    \includegraphics[width=\linewidth]{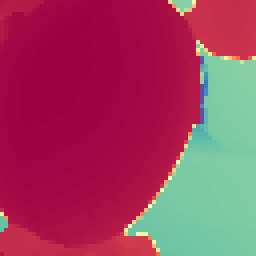}\vspace{\FigGridVspace} 
    \includegraphics[width=\linewidth]{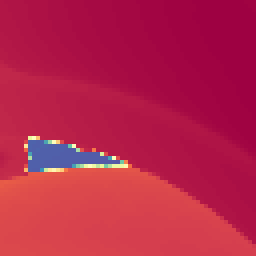}\vspace{\FigGridVspace}\vspace{\FigGridVspacee}
    
    \includegraphics[width=\linewidth]{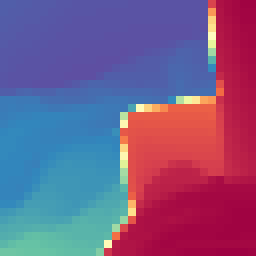}\vspace{\FigGridVspace} 
    \includegraphics[width=\linewidth]{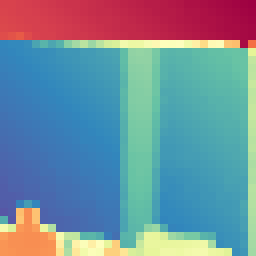}\vspace{\FigGridVspace}\vspace{\FigGridVspacee} 
    
    \includegraphics[width=\linewidth]{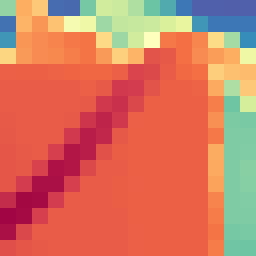}\vspace{\FigGridVspace}
    \includegraphics[width=\linewidth]{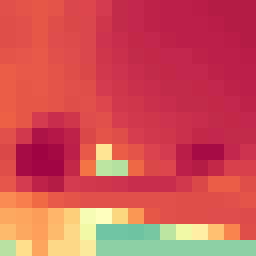}\vspace{\FigGridVspace}
    \end{minipage}

}\hspace{\FigGridHspace}
\subfloat[{GT}.]{
    \begin{minipage}[b]{\FigGridWidth\linewidth} 
    \includegraphics[width=\linewidth]{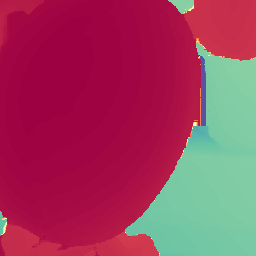}\vspace{\FigGridVspace}
    \includegraphics[width=\linewidth]{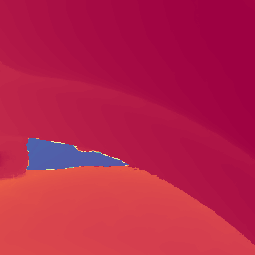}\vspace{\FigGridVspace}\vspace{\FigGridVspacee}    
    
    \includegraphics[width=\linewidth]{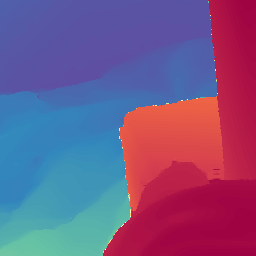}\vspace{\FigGridVspace}
    \includegraphics[width=\linewidth]{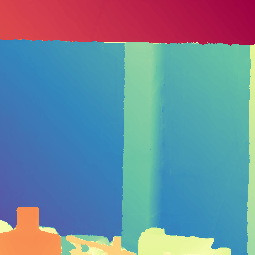}\vspace{\FigGridVspace}\vspace{\FigGridVspacee}
    
    \includegraphics[width=\linewidth]{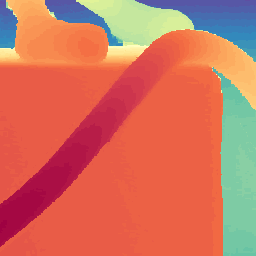}\vspace{\FigGridVspace}
    \includegraphics[width=\linewidth]{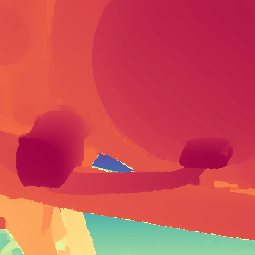}\vspace{\FigGridVspace}
    \end{minipage}
}\hspace{\FigGridHspace}\hspace{\FigGridHspacee}
\subfloat[{SD}.]{
    \begin{minipage}[b]{\FigGridWidth\linewidth} 
    \includegraphics[width=\linewidth]{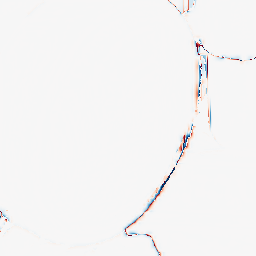}\vspace{\FigGridVspace}
    \includegraphics[width=\linewidth]{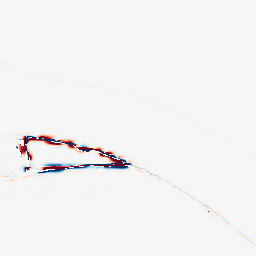}\vspace{\FigGridVspace}\vspace{\FigGridVspacee}
        
    \includegraphics[width=\linewidth]{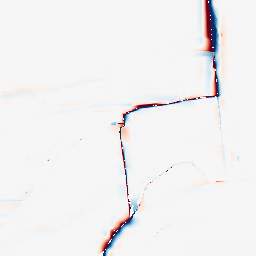}\vspace{\FigGridVspace}
    \includegraphics[width=\linewidth]{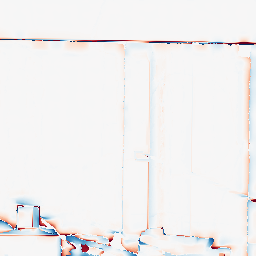}\vspace{\FigGridVspace}\vspace{\FigGridVspacee}
    
    \includegraphics[width=\linewidth]{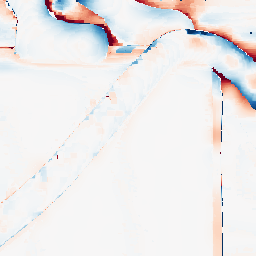}\vspace{\FigGridVspace}
    \includegraphics[width=\linewidth]{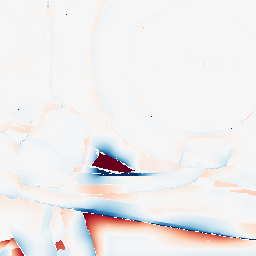}\vspace{\FigGridVspace}
    \end{minipage}
}\hspace{\FigGridHspace}
\subfloat[{Pixtrans}.]{
    \begin{minipage}[b]{\FigGridWidth\linewidth} 
    \includegraphics[width=\linewidth]{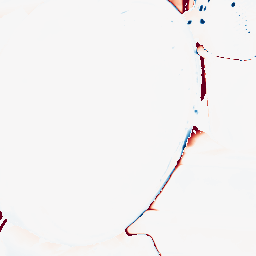}\vspace{\FigGridVspace}
    \includegraphics[width=\linewidth]{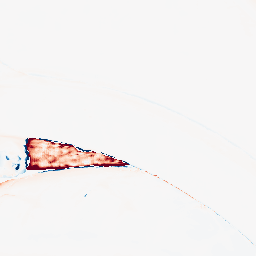}\vspace{\FigGridVspace}\vspace{\FigGridVspacee}  
    
    \includegraphics[width=\linewidth]{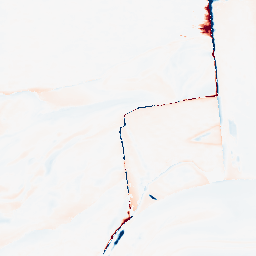}\vspace{\FigGridVspace}
    \includegraphics[width=\linewidth]{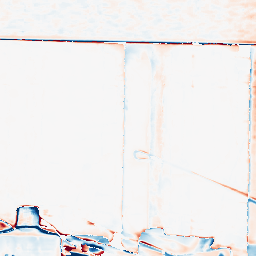}\vspace{\FigGridVspace}\vspace{\FigGridVspacee}  
    
    \includegraphics[width=\linewidth]{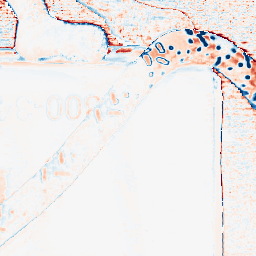}\vspace{\FigGridVspace}
    \includegraphics[width=\linewidth]{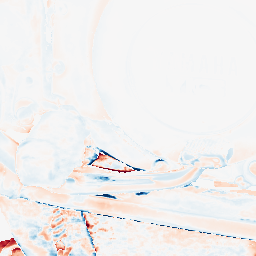}\vspace{\FigGridVspace}
    \end{minipage}
}\hspace{\FigGridHspace}
\subfloat[{MSGNet}.]{
    \begin{minipage}[b]{\FigGridWidth\linewidth} 
    \includegraphics[width=\linewidth]{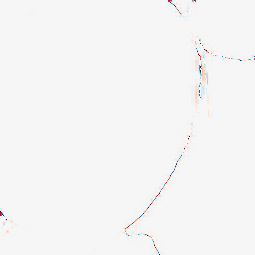}\vspace{\FigGridVspace}
    \includegraphics[width=\linewidth]{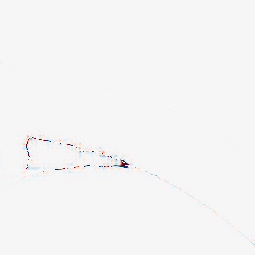}\vspace{\FigGridVspace}\vspace{\FigGridVspacee}
    
    \includegraphics[width=\linewidth]{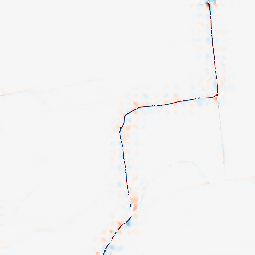}\vspace{\FigGridVspace}
    \includegraphics[width=\linewidth]{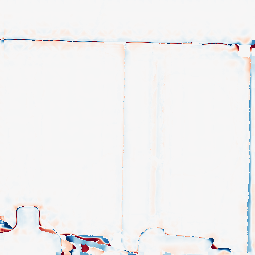}\vspace{\FigGridVspace}\vspace{\FigGridVspacee}
    
    \includegraphics[width=\linewidth]{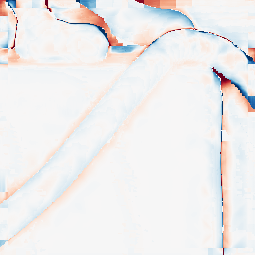}\vspace{\FigGridVspace}
    \includegraphics[width=\linewidth]{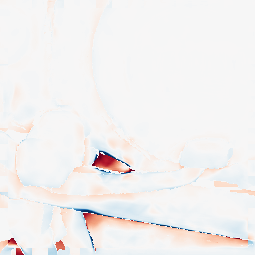}\vspace{\FigGridVspace}

    \end{minipage}
}\hspace{\FigGridHspace}
\subfloat[{FDKN}.]{
    \begin{minipage}[b]{\FigGridWidth\linewidth} 
    \includegraphics[width=\linewidth]{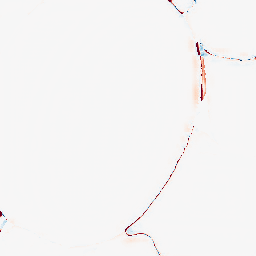}\vspace{\FigGridVspace}
    \includegraphics[width=\linewidth]{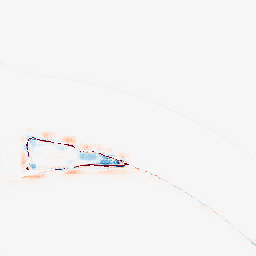}\vspace{\FigGridVspace}\vspace{\FigGridVspacee}

    \includegraphics[width=\linewidth]{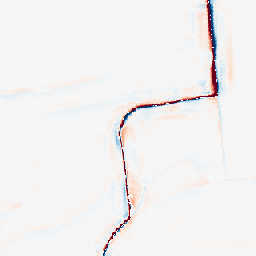}\vspace{\FigGridVspace}
    \includegraphics[width=\linewidth]{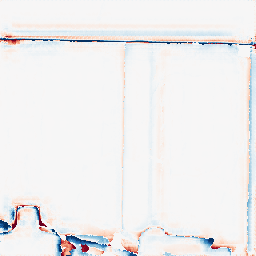}\vspace{\FigGridVspace}\vspace{\FigGridVspacee}
    
    \includegraphics[width=\linewidth]{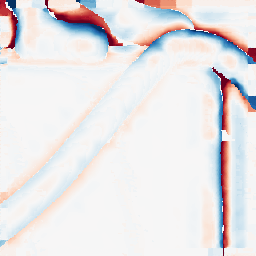}\vspace{\FigGridVspace}
    \includegraphics[width=\linewidth]{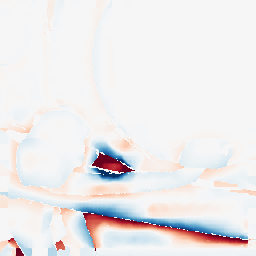}\vspace{\FigGridVspace}
    \end{minipage}
}\hspace{\FigGridHspace}
\subfloat[{PMBA}.]{
    \begin{minipage}[b]{\FigGridWidth\linewidth} 
    \includegraphics[width=\linewidth]{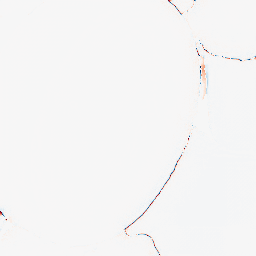}\vspace{\FigGridVspace}
    \includegraphics[width=\linewidth]{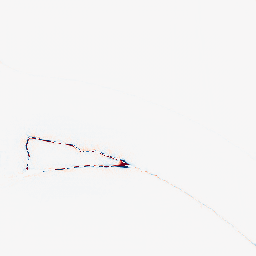}\vspace{\FigGridVspace}\vspace{\FigGridVspacee}
    
    \includegraphics[width=\linewidth]{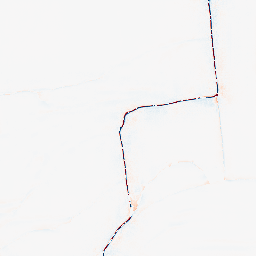}\vspace{\FigGridVspace}
    \includegraphics[width=\linewidth]{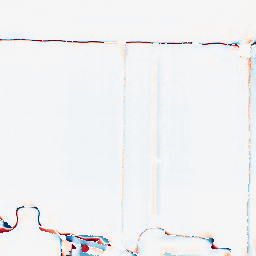}\vspace{\FigGridVspace}\vspace{\FigGridVspacee}
    
    \includegraphics[width=\linewidth]{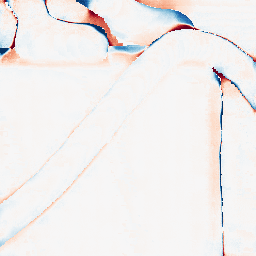}\vspace{\FigGridVspace}
    \includegraphics[width=\linewidth]{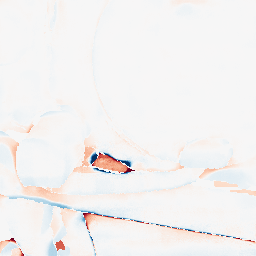}\vspace{\FigGridVspace}
    \end{minipage}
}\hspace{\FigGridHspace}
\subfloat[{FDSR}.]{
    \begin{minipage}[b]{\FigGridWidth\linewidth} 
    \includegraphics[width=\linewidth]{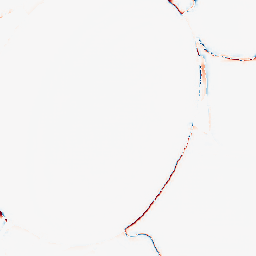}\vspace{\FigGridVspace}
    \includegraphics[width=\linewidth]{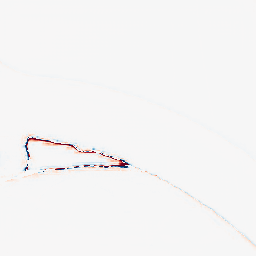}\vspace{\FigGridVspace}\vspace{\FigGridVspacee}
    
    \includegraphics[width=\linewidth]{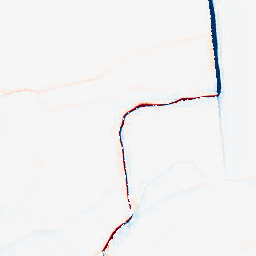}\vspace{\FigGridVspace}
    \includegraphics[width=\linewidth]{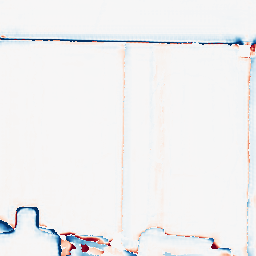}\vspace{\FigGridVspace}\vspace{\FigGridVspacee}

    \includegraphics[width=\linewidth]{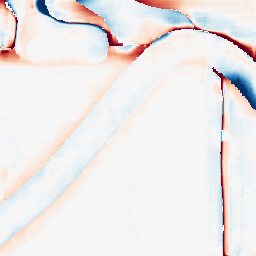}\vspace{\FigGridVspace}
    \includegraphics[width=\linewidth]{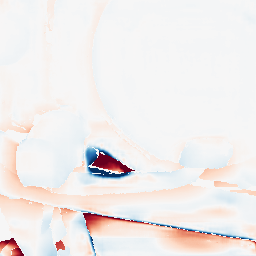}\vspace{\FigGridVspace}
    \end{minipage}
}\hspace{\FigGridHspace}\hspace{\FigGridHspacee}
\subfloat[{Ours}.]{
    \begin{minipage}[b]{\FigGridWidth\linewidth} 
    \includegraphics[width=\linewidth]{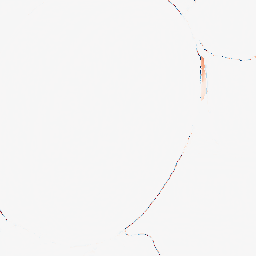}\vspace{\FigGridVspace}
     \includegraphics[width=\linewidth]{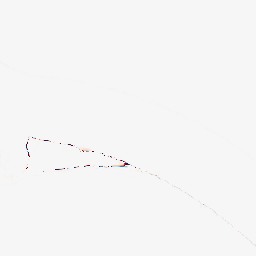}\vspace{\FigGridVspace}\vspace{\FigGridVspacee}
     
    \includegraphics[width=\linewidth]{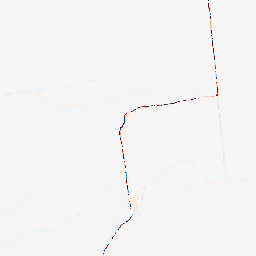}\vspace{\FigGridVspace}
    \includegraphics[width=\linewidth]{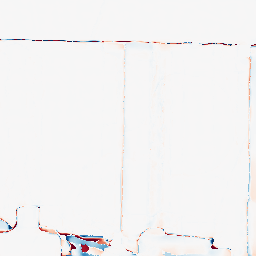}\vspace{\FigGridVspace}\vspace{\FigGridVspacee}
    
    \includegraphics[width=\linewidth]{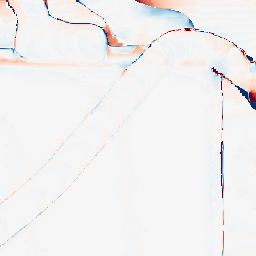}\vspace{\FigGridVspace}
    \includegraphics[width=\linewidth]{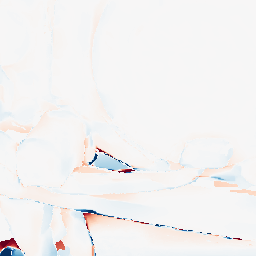}\vspace{\FigGridVspace}
  
    \end{minipage}
}\hspace{\FigGridHspace}
\subfloat[{Ours Pred}.]{
    \begin{minipage}[b]{\FigGridWidth\linewidth} 
    \includegraphics[width=\linewidth]{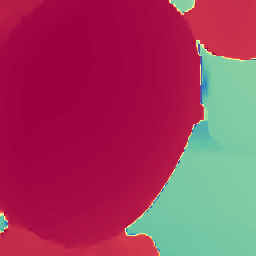}\vspace{\FigGridVspace}
    \includegraphics[width=\linewidth]{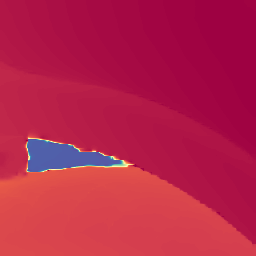}\vspace{\FigGridVspace}\vspace{\FigGridVspacee}
    
    \includegraphics[width=\linewidth]{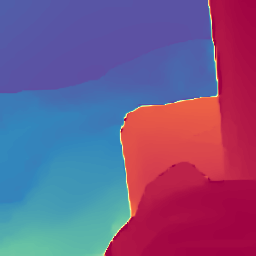}\vspace{\FigGridVspace}
    \includegraphics[width=\linewidth]{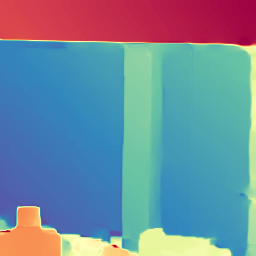}\vspace{\FigGridVspace}\vspace{\FigGridVspacee}
    
    \includegraphics[width=\linewidth]{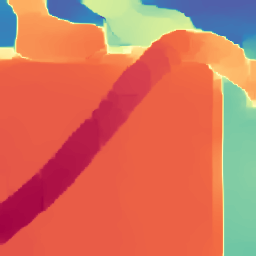}\vspace{\FigGridVspace}
    \includegraphics[width=\linewidth]{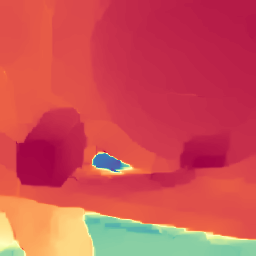}\vspace{\FigGridVspace}
    \end{minipage}
}
\caption{Additional qualitative comparison of upsampled depth maps for the Middlebury dataset \cite{Scharstein2007,Scharstein2001,Scharstein2014,Scharstein2003,Hirschmuller2007}. From top to bottom each group of two rows shows the error of upsampled images, defined as the difference between the prediction and the ground truth, for upsampling factors $\times4$, $\times8$ and $\times16$ respectively. From left to right, the first group of columns are (a) Guide, (b) Source and (c) Ground Truth; the second group includes selected methods from our quantitative evaluation, (d) SDF \cite{Ham2018}, (e) Pixtransform \cite{deLutio2019}, (f) MSGNet \cite{Tak-Wai2016}, (g) FDKN \cite{Kim2021}, (h) PMBANet \cite{Ye2020} and (i) FDSR \cite{he2021}; the last two columns represent (j) the error for the prediction of our model and (k) the prediction itself.}
\label{sup_fig:depth_upsampling_middlebury}
\vspace{-0.3cm}
\end{sidewaysfigure*}
\newcommand\NYU{201}
\newcommand\NYUeight{97}
\newcommand\NYUsixteen{35}

\newcommand\bisNYU{421}
\newcommand\bisNYUeight{210}
\newcommand\bisNYUsixteen{496}

\begin{sidewaysfigure*}[p]
\vspace{-0.2cm}
  \centering
  \tiny	
\subfloat[{Guide.}]{
    \begin{minipage}[b]{\FigGridWidth\linewidth} 
        \includegraphics[width=\linewidth]{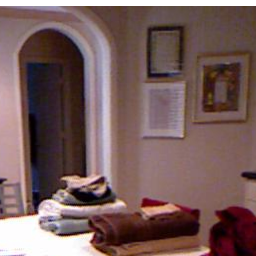}\vspace{\FigGridVspace}         
        \includegraphics[width=\linewidth]{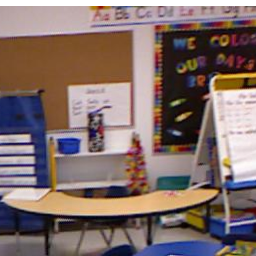}\vspace{\FigGridVspace}\vspace{\FigGridVspacee}        
        
        \includegraphics[width=\linewidth]{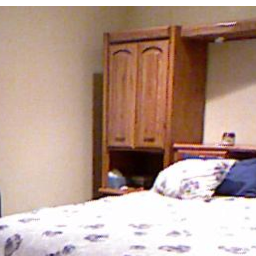}\vspace{\FigGridVspace} 
        \includegraphics[width=\linewidth]{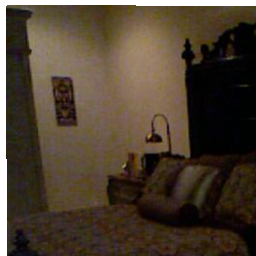}\vspace{\FigGridVspace}\vspace{\FigGridVspacee}   
   
        \includegraphics[width=\linewidth]{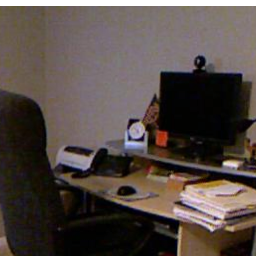}\vspace{\FigGridVspace}
        \includegraphics[width=\linewidth]{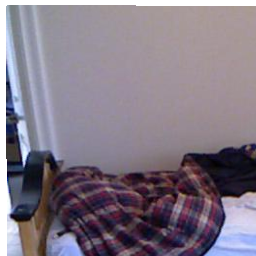}\vspace{\FigGridVspace}
    \end{minipage}
}\hspace{\FigGridHspace}
\subfloat[{Source}.]{
    \begin{minipage}[b]{\FigGridWidth\linewidth} 
    \includegraphics[width=\linewidth]{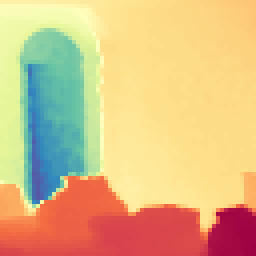}\vspace{\FigGridVspace} 
    \includegraphics[width=\linewidth]{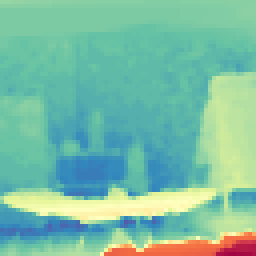}\vspace{\FigGridVspace}\vspace{\FigGridVspacee}
    
    \includegraphics[width=\linewidth]{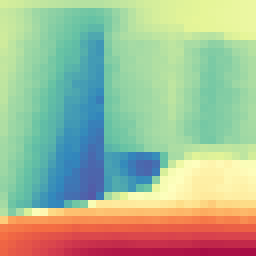}\vspace{\FigGridVspace} 
    \includegraphics[width=\linewidth]{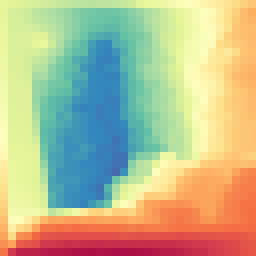}\vspace{\FigGridVspace}\vspace{\FigGridVspacee} 
    
    \includegraphics[width=\linewidth]{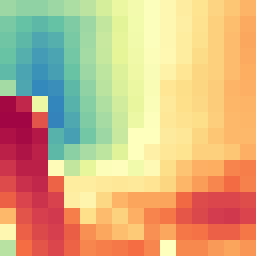}\vspace{\FigGridVspace}
    \includegraphics[width=\linewidth]{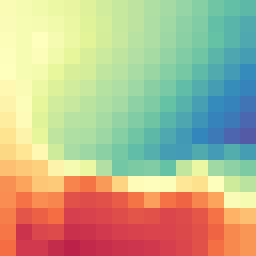}\vspace{\FigGridVspace}
    \end{minipage}

}\hspace{\FigGridHspace}
\subfloat[{GT}.]{
    \begin{minipage}[b]{\FigGridWidth\linewidth} 
    \includegraphics[width=\linewidth]{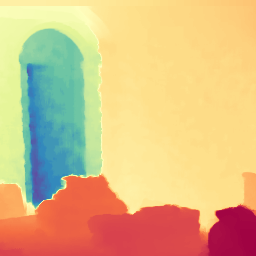}\vspace{\FigGridVspace}
    \includegraphics[width=\linewidth]{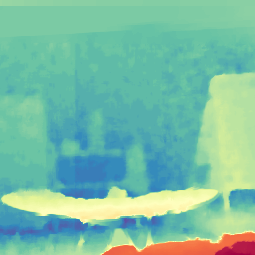}\vspace{\FigGridVspace}\vspace{\FigGridVspacee}    
    
    \includegraphics[width=\linewidth]{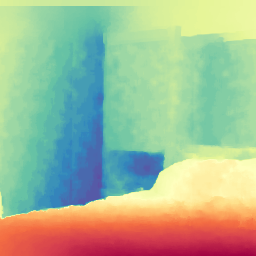}\vspace{\FigGridVspace}
    \includegraphics[width=\linewidth]{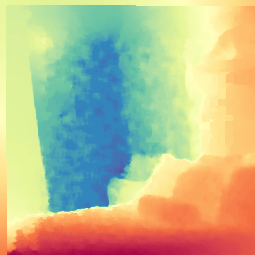}\vspace{\FigGridVspace}\vspace{\FigGridVspacee}
    
    \includegraphics[width=\linewidth]{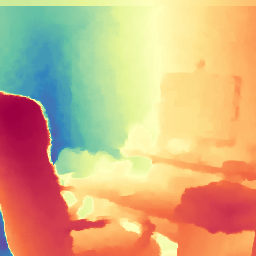}\vspace{\FigGridVspace}
    \includegraphics[width=\linewidth]{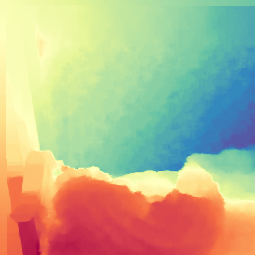}\vspace{\FigGridVspace}
    \end{minipage}
}\hspace{\FigGridHspace}\hspace{\FigGridHspacee}
\subfloat[{SD}.]{
    \begin{minipage}[b]{\FigGridWidth\linewidth} 
    \includegraphics[width=\linewidth]{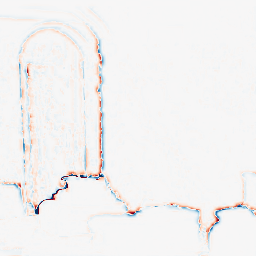}\vspace{\FigGridVspace}
    \includegraphics[width=\linewidth]{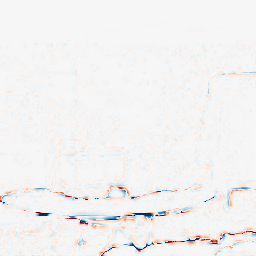}\vspace{\FigGridVspace}\vspace{\FigGridVspacee}
        
    \includegraphics[width=\linewidth]{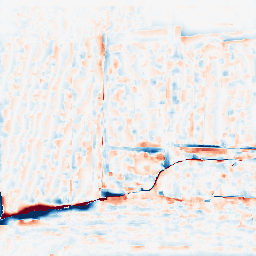}\vspace{\FigGridVspace}
    \includegraphics[width=\linewidth]{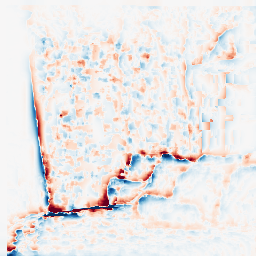}\vspace{\FigGridVspace}\vspace{\FigGridVspacee}
    
    \includegraphics[width=\linewidth]{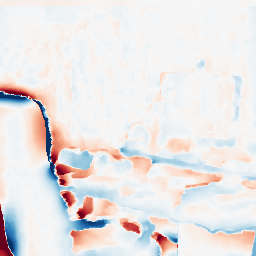}\vspace{\FigGridVspace}
    \includegraphics[width=\linewidth]{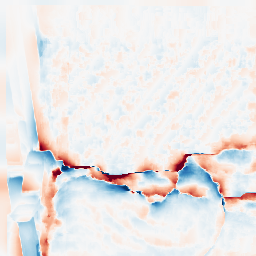}\vspace{\FigGridVspace}
    \end{minipage}
}\hspace{\FigGridHspace}
\subfloat[{Pixtrans}.]{
    \begin{minipage}[b]{\FigGridWidth\linewidth} 
    \includegraphics[width=\linewidth]{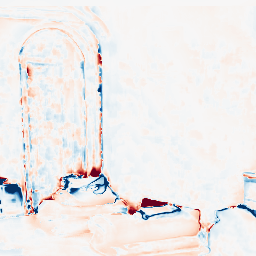}\vspace{\FigGridVspace}
    \includegraphics[width=\linewidth]{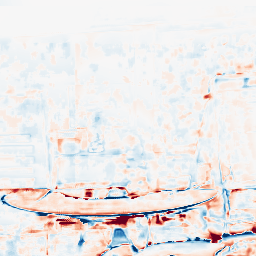}\vspace{\FigGridVspace}\vspace{\FigGridVspacee}  
    
    \includegraphics[width=\linewidth]{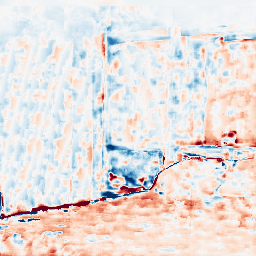}\vspace{\FigGridVspace}
    \includegraphics[width=\linewidth]{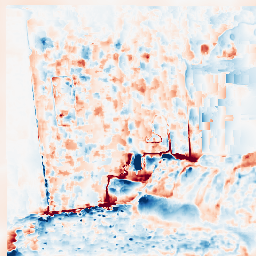}\vspace{\FigGridVspace}\vspace{\FigGridVspacee}  
    
    \includegraphics[width=\linewidth]{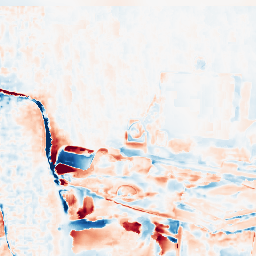}\vspace{\FigGridVspace}
    \includegraphics[width=\linewidth]{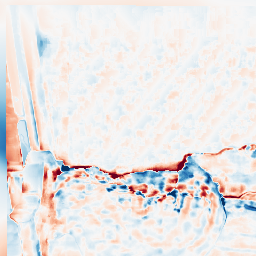}\vspace{\FigGridVspace}
    \end{minipage}
}\hspace{\FigGridHspace}
\subfloat[{MSGNet}.]{
    \begin{minipage}[b]{\FigGridWidth\linewidth} 
    \includegraphics[width=\linewidth]{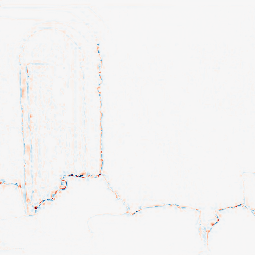}\vspace{\FigGridVspace}
    \includegraphics[width=\linewidth]{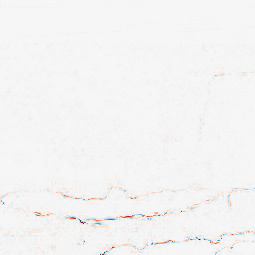}\vspace{\FigGridVspace}\vspace{\FigGridVspacee}
    
    \includegraphics[width=\linewidth]{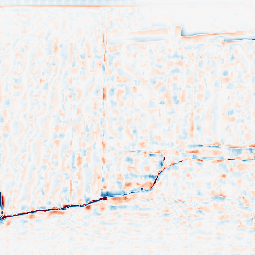}\vspace{\FigGridVspace}
    \includegraphics[width=\linewidth]{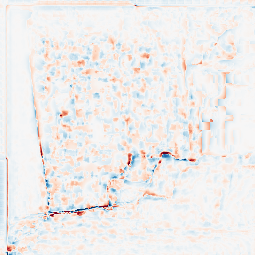}\vspace{\FigGridVspace}\vspace{\FigGridVspacee}
    
    \includegraphics[width=\linewidth]{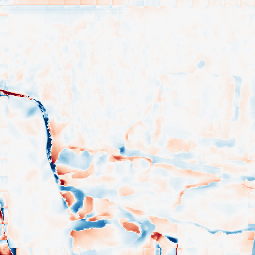}\vspace{\FigGridVspace}
    \includegraphics[width=\linewidth]{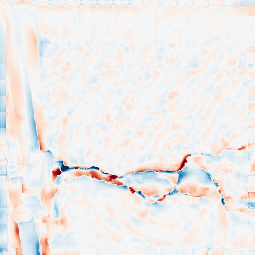}\vspace{\FigGridVspace}

    \end{minipage}
}\hspace{\FigGridHspace}
\subfloat[{FDKN}.]{
    \begin{minipage}[b]{\FigGridWidth\linewidth} 
    \includegraphics[width=\linewidth]{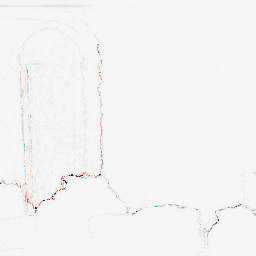}\vspace{\FigGridVspace}
    \includegraphics[width=\linewidth]{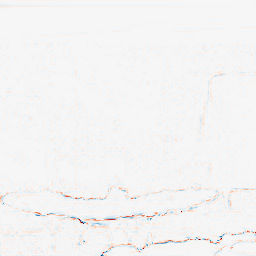}\vspace{\FigGridVspace}\vspace{\FigGridVspacee}

    \includegraphics[width=\linewidth]{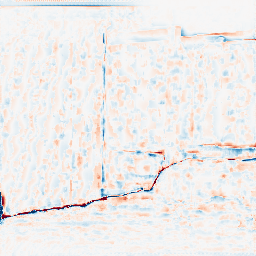}\vspace{\FigGridVspace}
    \includegraphics[width=\linewidth]{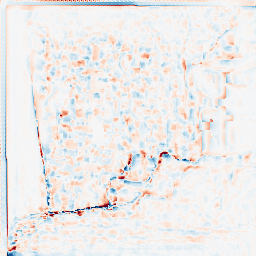}\vspace{\FigGridVspace}\vspace{\FigGridVspacee}
    
    \includegraphics[width=\linewidth]{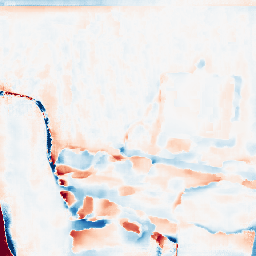}\vspace{\FigGridVspace}
    \includegraphics[width=\linewidth]{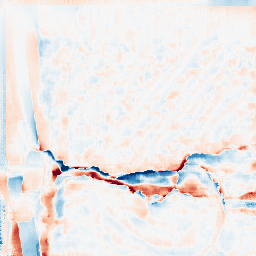}\vspace{\FigGridVspace}
    \end{minipage}
}\hspace{\FigGridHspace}
\subfloat[{PMBA}.]{
    \begin{minipage}[b]{\FigGridWidth\linewidth} 
    \includegraphics[width=\linewidth]{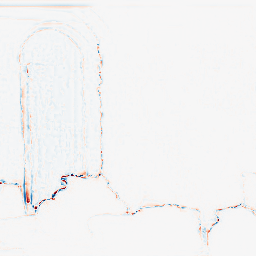}\vspace{\FigGridVspace}
    \includegraphics[width=\linewidth]{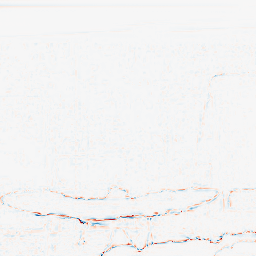}\vspace{\FigGridVspace}\vspace{\FigGridVspacee}
    
    \includegraphics[width=\linewidth]{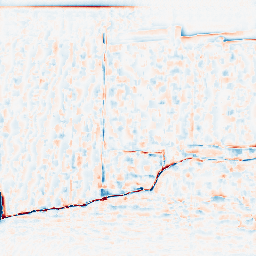}\vspace{\FigGridVspace}
    \includegraphics[width=\linewidth]{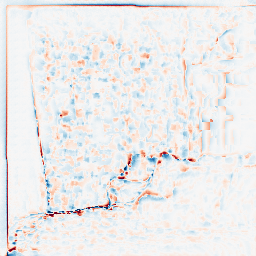}\vspace{\FigGridVspace}\vspace{\FigGridVspacee}
    
    \includegraphics[width=\linewidth]{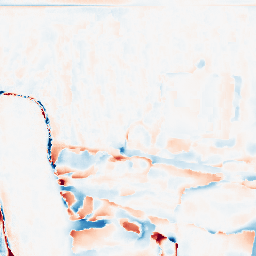}\vspace{\FigGridVspace}
    \includegraphics[width=\linewidth]{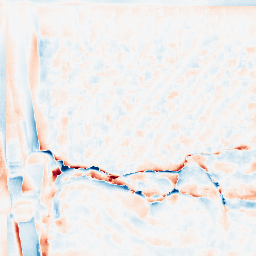}\vspace{\FigGridVspace}
    \end{minipage}
}\hspace{\FigGridHspace}
\subfloat[{FDSR}.]{
    \begin{minipage}[b]{\FigGridWidth\linewidth} 
    \includegraphics[width=\linewidth]{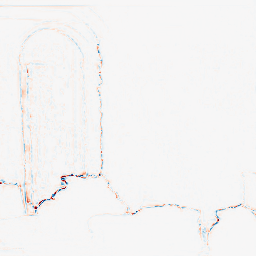}\vspace{\FigGridVspace}
    \includegraphics[width=\linewidth]{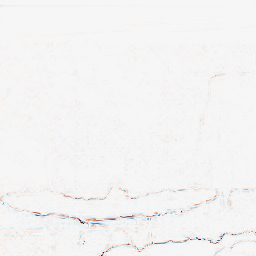}\vspace{\FigGridVspace}\vspace{\FigGridVspacee}
    
    \includegraphics[width=\linewidth]{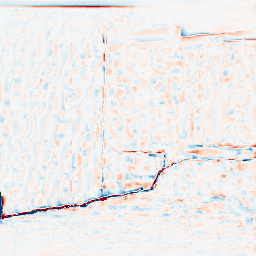}\vspace{\FigGridVspace}
    \includegraphics[width=\linewidth]{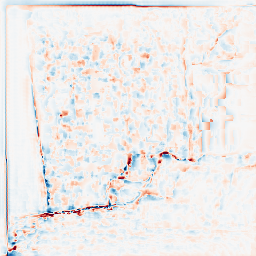}\vspace{\FigGridVspace}\vspace{\FigGridVspacee}

    \includegraphics[width=\linewidth]{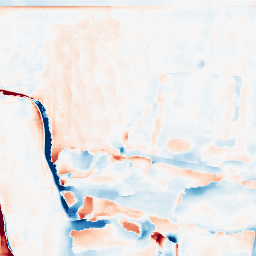}\vspace{\FigGridVspace}
    \includegraphics[width=\linewidth]{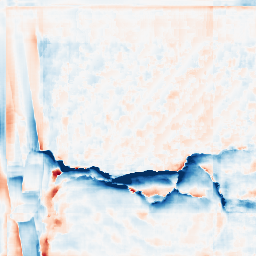}\vspace{\FigGridVspace}
    \end{minipage}
}\hspace{\FigGridHspace}\hspace{\FigGridHspacee}
\subfloat[{Ours}.]{
    \begin{minipage}[b]{\FigGridWidth\linewidth} 
    \includegraphics[width=\linewidth]{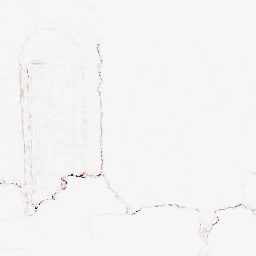}\vspace{\FigGridVspace}
     \includegraphics[width=\linewidth]{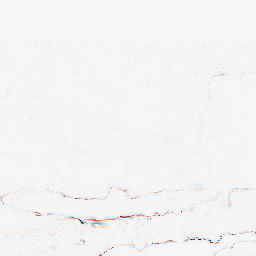}\vspace{\FigGridVspace}\vspace{\FigGridVspacee}
     
    \includegraphics[width=\linewidth]{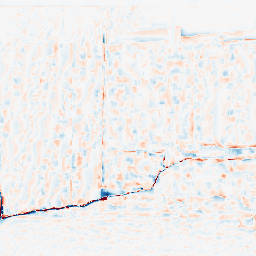}\vspace{\FigGridVspace}
    \includegraphics[width=\linewidth]{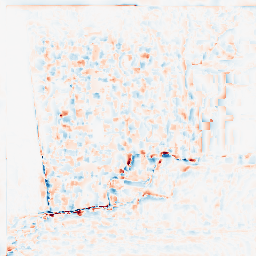}\vspace{\FigGridVspace}\vspace{\FigGridVspacee}
    
    \includegraphics[width=\linewidth]{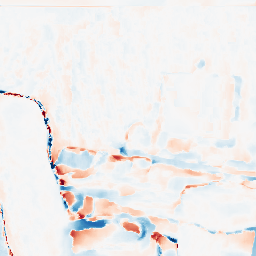}\vspace{\FigGridVspace}
    \includegraphics[width=\linewidth]{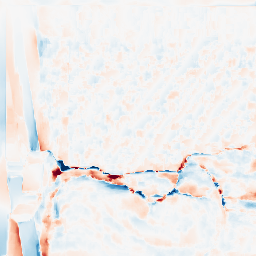}\vspace{\FigGridVspace}
  
    \end{minipage}
}\hspace{\FigGridHspace}
\subfloat[{Ours Pred}.]{
    \begin{minipage}[b]{\FigGridWidth\linewidth} 
    \includegraphics[width=\linewidth]{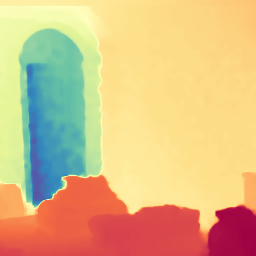}\vspace{\FigGridVspace}
    \includegraphics[width=\linewidth]{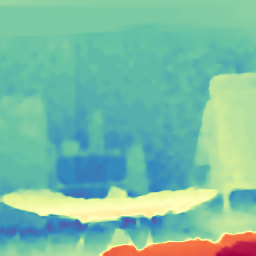}\vspace{\FigGridVspace}\vspace{\FigGridVspacee}
    
    \includegraphics[width=\linewidth]{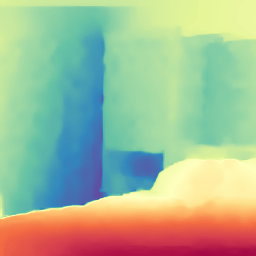}\vspace{\FigGridVspace}
    \includegraphics[width=\linewidth]{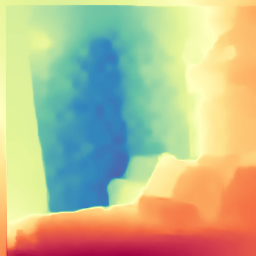}\vspace{\FigGridVspace}\vspace{\FigGridVspacee}
    
    \includegraphics[width=\linewidth]{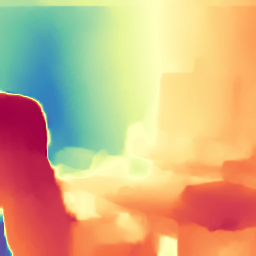}\vspace{\FigGridVspace}
    \includegraphics[width=\linewidth]{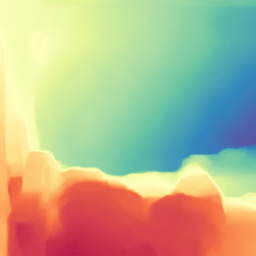}\vspace{\FigGridVspace}
    \end{minipage}
}
\caption{Additional qualitative comparison of upsampled depth maps for the NYUv2 dataset \cite{Silberman2012}. From top to bottom each group of two rows shows the error of upsampled images, defined as the difference between the prediction and the ground truth, for upsampling factors $\times4$, $\times8$ and $\times16$ respectively. From left to right, the first group of columns are (a) Guide, (b) Source and (c) Ground Truth; the second group includes selected methods from our quantitative evaluation, (d) SDF \cite{Ham2018}, (e) Pixtransform \cite{deLutio2019}, (f) MSGNet \cite{Tak-Wai2016}, (g) FDKN \cite{Kim2021}, (h) PMBANet \cite{Ye2020} and (i) FDSR \cite{he2021}; the last two columns represent (j) the error for the prediction of our model and (k) the prediction itself.}
\label{sup_fig:depth_upsampling_nyu}
\vspace{-0.3cm}
\end{sidewaysfigure*}
\newcommand\DIML{1310}
\newcommand\DIMLeight{3793}
\newcommand\DIMLsixteen{4755}

\newcommand\bisDIML{2616}
\newcommand\bisDIMLeight{2607}
\newcommand\bisDIMLsixteen{1806}

\begin{sidewaysfigure*}[p]
\vspace{-0.2cm}
  \centering
  \tiny	

\subfloat[{Guide.}]{
    \begin{minipage}[b]{\FigGridWidth\linewidth} 
        \includegraphics[width=\linewidth]{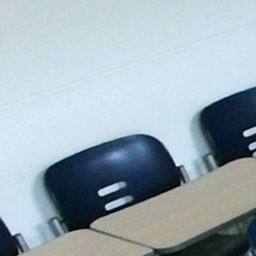}\vspace{\FigGridVspace}         
        \includegraphics[width=\linewidth]{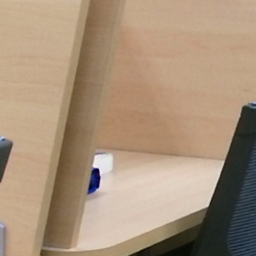}\vspace{\FigGridVspace}\vspace{\FigGridVspacee}        
        
        \includegraphics[width=\linewidth]{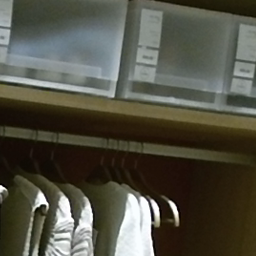}\vspace{\FigGridVspace} 
        \includegraphics[width=\linewidth]{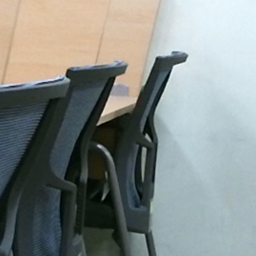}\vspace{\FigGridVspace}\vspace{\FigGridVspacee}   
   
        \includegraphics[width=\linewidth]{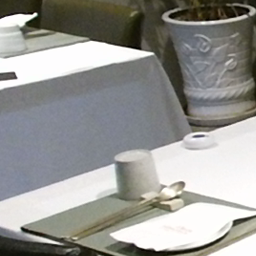}\vspace{\FigGridVspace}
        \includegraphics[width=\linewidth]{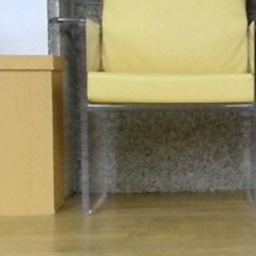}\vspace{\FigGridVspace}

    \end{minipage}
}\hspace{\FigGridHspace}
\subfloat[{Source}.]{

    \begin{minipage}[b]{\FigGridWidth\linewidth} 
    \includegraphics[width=\linewidth]{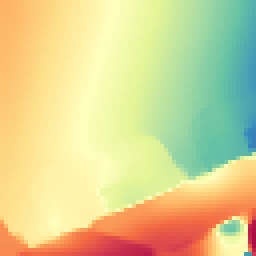}\vspace{\FigGridVspace} 
    \includegraphics[width=\linewidth]{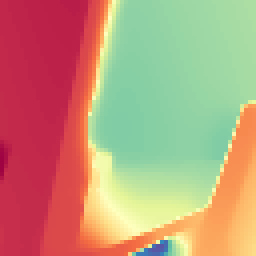}\vspace{\FigGridVspace}\vspace{\FigGridVspacee}
    
    \includegraphics[width=\linewidth]{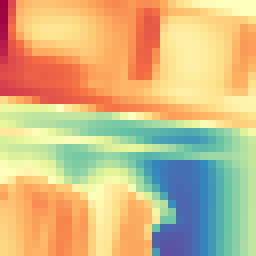}\vspace{\FigGridVspace} 
    \includegraphics[width=\linewidth]{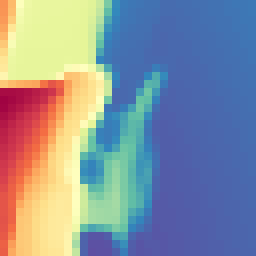}\vspace{\FigGridVspace}\vspace{\FigGridVspacee} 
    
    \includegraphics[width=\linewidth]{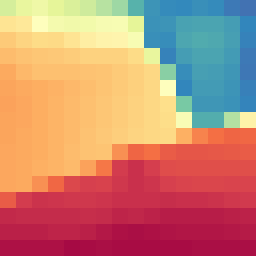}\vspace{\FigGridVspace}
    \includegraphics[width=\linewidth]{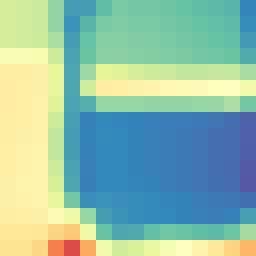}\vspace{\FigGridVspace}
    \end{minipage}

}\hspace{\FigGridHspace}
\subfloat[{GT}.]{
    \begin{minipage}[b]{\FigGridWidth\linewidth} 
    \includegraphics[width=\linewidth]{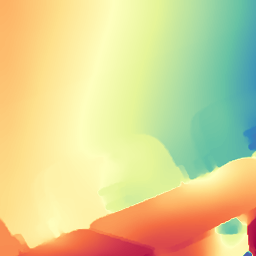}\vspace{\FigGridVspace}
    \includegraphics[width=\linewidth]{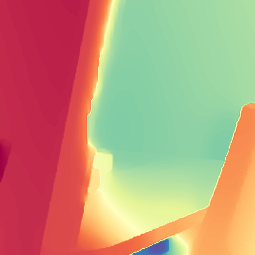}\vspace{\FigGridVspace}\vspace{\FigGridVspacee}    
    
    \includegraphics[width=\linewidth]{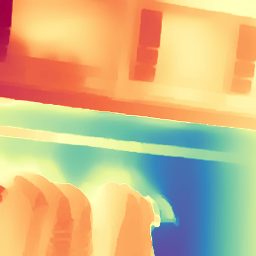}\vspace{\FigGridVspace}
    \includegraphics[width=\linewidth]{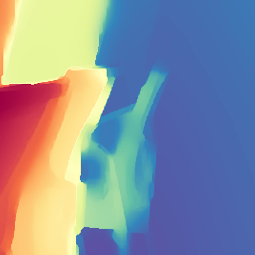}\vspace{\FigGridVspace}\vspace{\FigGridVspacee}
    
    \includegraphics[width=\linewidth]{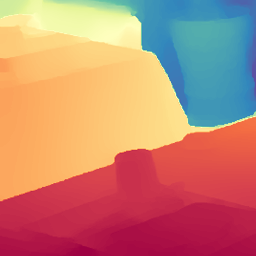}\vspace{\FigGridVspace}
    \includegraphics[width=\linewidth]{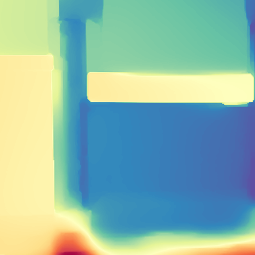}\vspace{\FigGridVspace}
    \end{minipage}
}\hspace{\FigGridHspace}\hspace{\FigGridHspacee}
\subfloat[{SD}.]{
    \begin{minipage}[b]{\FigGridWidth\linewidth} 
    \includegraphics[width=\linewidth]{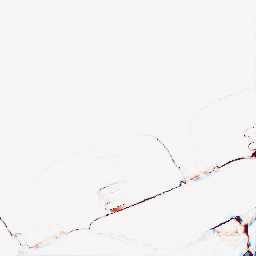}\vspace{\FigGridVspace}
    \includegraphics[width=\linewidth]{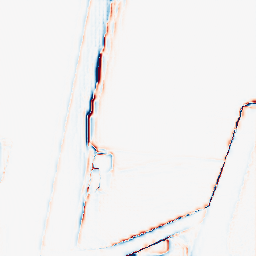}\vspace{\FigGridVspace}\vspace{\FigGridVspacee}
        
    \includegraphics[width=\linewidth]{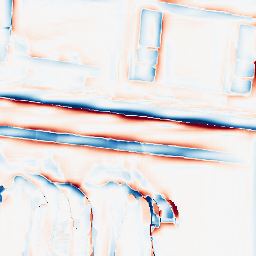}\vspace{\FigGridVspace}
    \includegraphics[width=\linewidth]{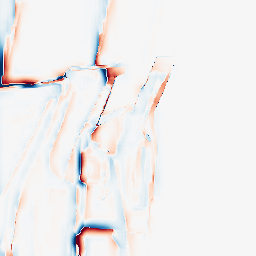}\vspace{\FigGridVspace}\vspace{\FigGridVspacee}
    
    \includegraphics[width=\linewidth]{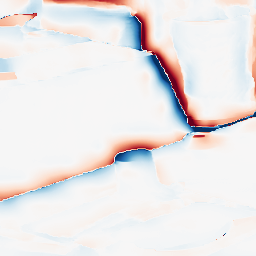}\vspace{\FigGridVspace}
    \includegraphics[width=\linewidth]{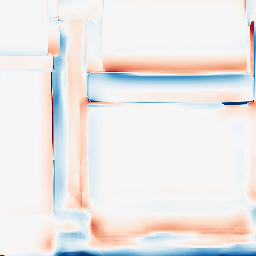}\vspace{\FigGridVspace}
    \end{minipage}
}\hspace{\FigGridHspace}
\subfloat[{Pixtrans}.]{
    \begin{minipage}[b]{\FigGridWidth\linewidth} 
    \includegraphics[width=\linewidth]{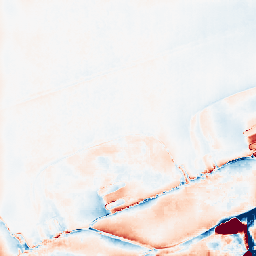}\vspace{\FigGridVspace}
    \includegraphics[width=\linewidth]{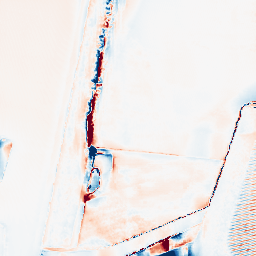}\vspace{\FigGridVspace}\vspace{\FigGridVspacee}  
    
    \includegraphics[width=\linewidth]{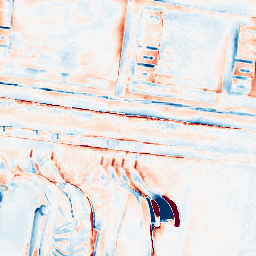}\vspace{\FigGridVspace}
    \includegraphics[width=\linewidth]{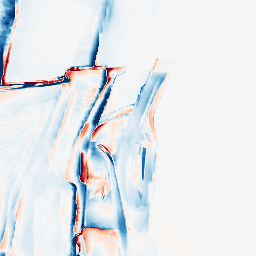}\vspace{\FigGridVspace}\vspace{\FigGridVspacee}  
    
    \includegraphics[width=\linewidth]{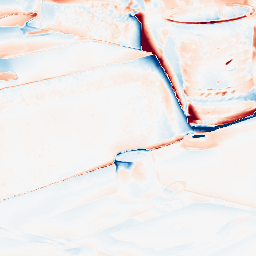}\vspace{\FigGridVspace}
    \includegraphics[width=\linewidth]{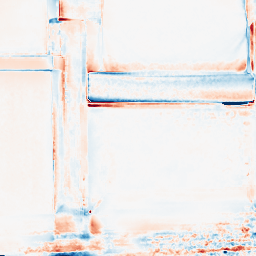}\vspace{\FigGridVspace}
    \end{minipage}
}\hspace{\FigGridHspace}
\subfloat[{MSGNet}.]{
    \begin{minipage}[b]{\FigGridWidth\linewidth} 
    \includegraphics[width=\linewidth]{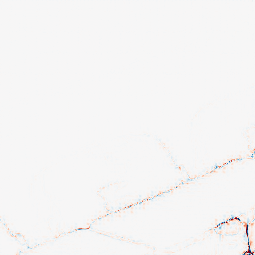}\vspace{\FigGridVspace}
    \includegraphics[width=\linewidth]{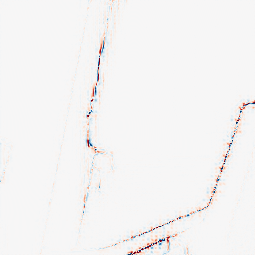}\vspace{\FigGridVspace}\vspace{\FigGridVspacee}
    
    \includegraphics[width=\linewidth]{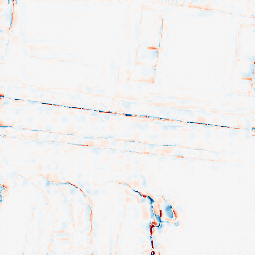}\vspace{\FigGridVspace}
    \includegraphics[width=\linewidth]{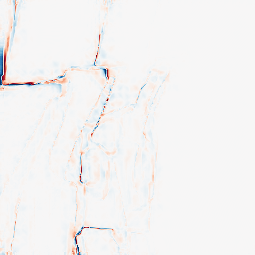}\vspace{\FigGridVspace}\vspace{\FigGridVspacee}
    
    \includegraphics[width=\linewidth]{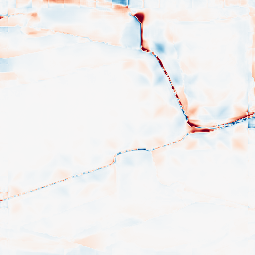}\vspace{\FigGridVspace}
    \includegraphics[width=\linewidth]{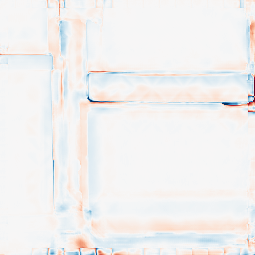}\vspace{\FigGridVspace}

    \end{minipage}
}\hspace{\FigGridHspace}
\subfloat[{FDKN}.]{
    \begin{minipage}[b]{\FigGridWidth\linewidth} 
    \includegraphics[width=\linewidth]{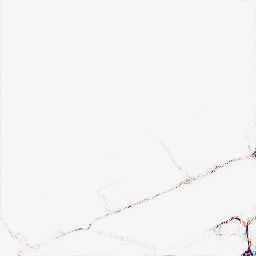}\vspace{\FigGridVspace}
    \includegraphics[width=\linewidth]{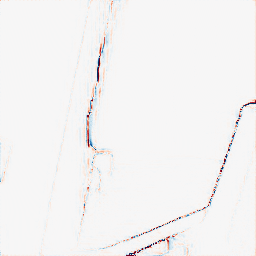}\vspace{\FigGridVspace}\vspace{\FigGridVspacee}

    \includegraphics[width=\linewidth]{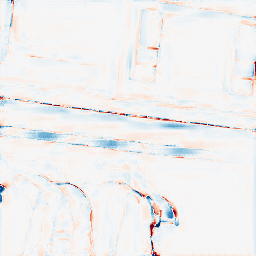}\vspace{\FigGridVspace}
    \includegraphics[width=\linewidth]{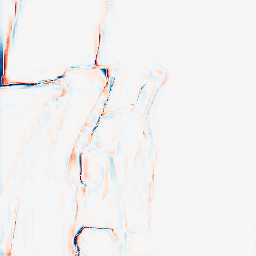}\vspace{\FigGridVspace}\vspace{\FigGridVspacee}
    
    \includegraphics[width=\linewidth]{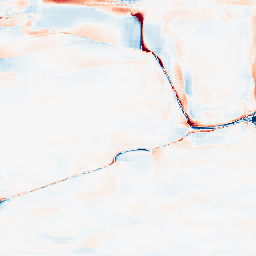}\vspace{\FigGridVspace}
    \includegraphics[width=\linewidth]{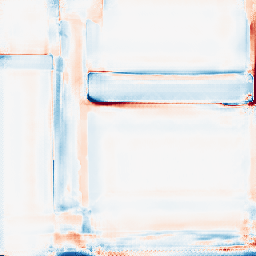}\vspace{\FigGridVspace}
    \end{minipage}
}\hspace{\FigGridHspace}
\subfloat[{PMBA}.]{
    \begin{minipage}[b]{\FigGridWidth\linewidth} 
    \includegraphics[width=\linewidth]{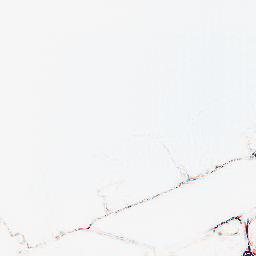}\vspace{\FigGridVspace}
    \includegraphics[width=\linewidth]{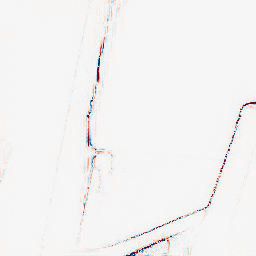}\vspace{\FigGridVspace}\vspace{\FigGridVspacee}
    
    \includegraphics[width=\linewidth]{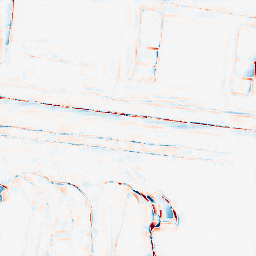}\vspace{\FigGridVspace}
    \includegraphics[width=\linewidth]{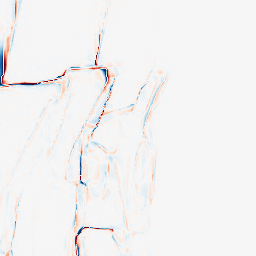}\vspace{\FigGridVspace}\vspace{\FigGridVspacee}
    
    \includegraphics[width=\linewidth]{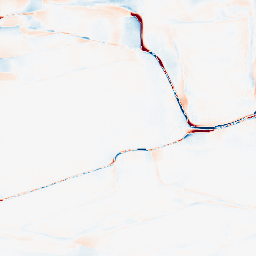}\vspace{\FigGridVspace}
    \includegraphics[width=\linewidth]{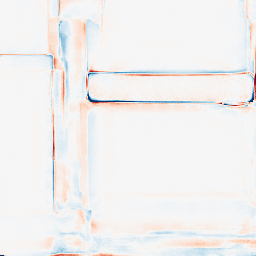}\vspace{\FigGridVspace}
    \end{minipage}
}\hspace{\FigGridHspace}
\subfloat[{FDSR}.]{
    \begin{minipage}[b]{\FigGridWidth\linewidth} 
    \includegraphics[width=\linewidth]{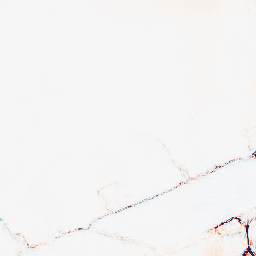}\vspace{\FigGridVspace}
    \includegraphics[width=\linewidth]{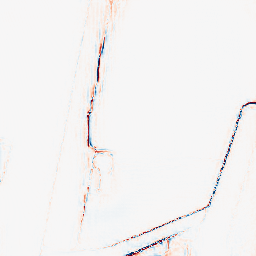}\vspace{\FigGridVspace}\vspace{\FigGridVspacee}
    
    \includegraphics[width=\linewidth]{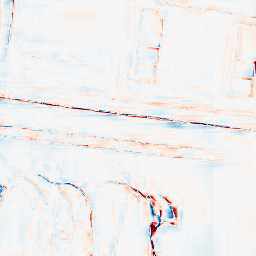}\vspace{\FigGridVspace}
    \includegraphics[width=\linewidth]{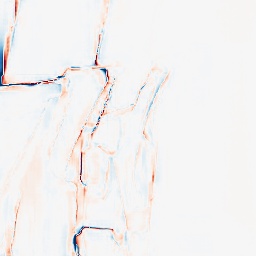}\vspace{\FigGridVspace}\vspace{\FigGridVspacee}

    \includegraphics[width=\linewidth]{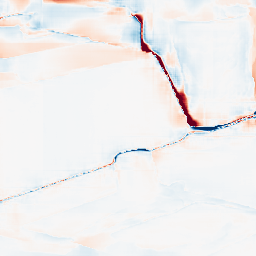}\vspace{\FigGridVspace}
    \includegraphics[width=\linewidth]{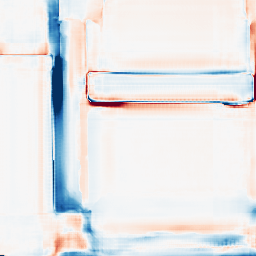}\vspace{\FigGridVspace}
    \end{minipage}
}\hspace{\FigGridHspace}\hspace{\FigGridHspacee}
\subfloat[{Ours}.]{
    \begin{minipage}[b]{\FigGridWidth\linewidth} 
    \includegraphics[width=\linewidth]{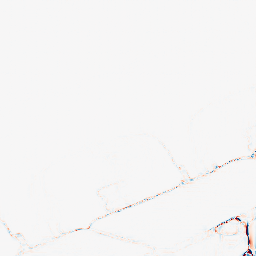}\vspace{\FigGridVspace}
     \includegraphics[width=\linewidth]{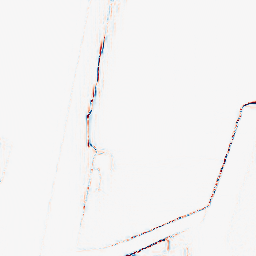}\vspace{\FigGridVspace}\vspace{\FigGridVspacee}
     
    \includegraphics[width=\linewidth]{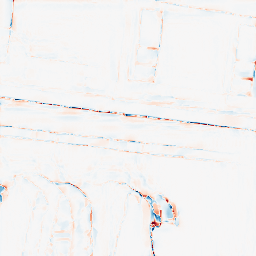}\vspace{\FigGridVspace}
    \includegraphics[width=\linewidth]{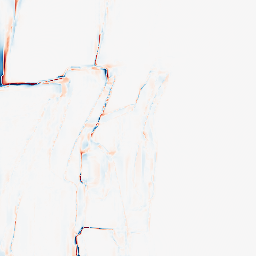}\vspace{\FigGridVspace}\vspace{\FigGridVspacee}
    
    \includegraphics[width=\linewidth]{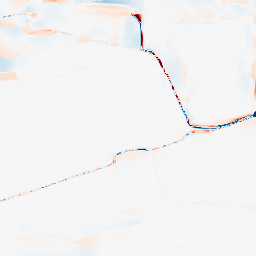}\vspace{\FigGridVspace}
    \includegraphics[width=\linewidth]{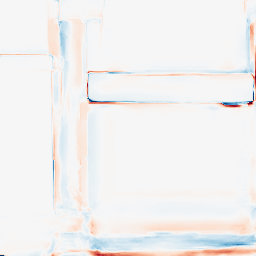}\vspace{\FigGridVspace}
  
    \end{minipage}
}\hspace{\FigGridHspace}
\subfloat[{Ours Pred}.]{
    \begin{minipage}[b]{\FigGridWidth\linewidth} 
    \includegraphics[width=\linewidth]{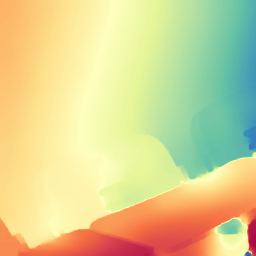}\vspace{\FigGridVspace}
    \includegraphics[width=\linewidth]{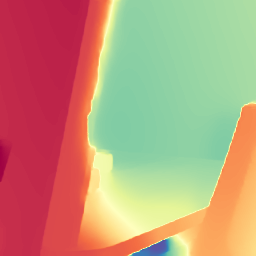}\vspace{\FigGridVspace}\vspace{\FigGridVspacee}
    
    \includegraphics[width=\linewidth]{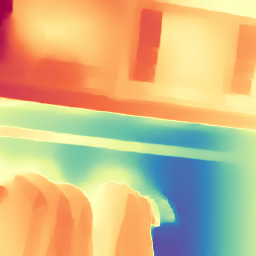}\vspace{\FigGridVspace}
    \includegraphics[width=\linewidth]{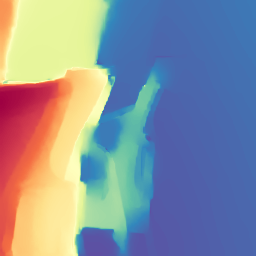}\vspace{\FigGridVspace}\vspace{\FigGridVspacee}
    
    \includegraphics[width=\linewidth]{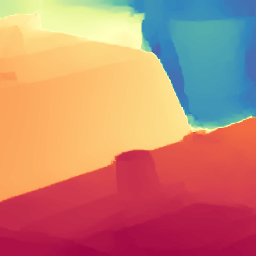}\vspace{\FigGridVspace}
    \includegraphics[width=\linewidth]{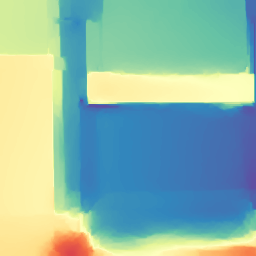}\vspace{\FigGridVspace}
    \end{minipage}
}
\caption{Additional qualitative comparison of upsampled depth maps for the DIML dataset \cite{DIML1,DIML2,DIML3,DIML4}. From top to bottom each group of two rows shows the error of upsampled images, defined as the difference between the prediction and the ground truth, for upsampling factors $\times4$, $\times8$ and $\times16$ respectively. From left to right, the first group of columns are (a) Guide, (b) Source and (c) Ground Truth; the second group includes selected methods from our quantitative evaluation, (d) SDF \cite{Ham2018}, (e) Pixtransform \cite{deLutio2019}, (f) MSGNet \cite{Tak-Wai2016}, (g) FDKN \cite{Kim2021}, (h) PMBANet \cite{Ye2020} and (i) FDSR \cite{he2021}; the last two columns represent (j) the error for the prediction of our model and (k) the prediction itself.}
\label{sup_fig:depth_upsampling_diml}
\vspace{-0.3cm}
\end{sidewaysfigure*}

\end{document}